\documentclass[10pt,journal,compsoc]{IEEEtran}
\usepackage{amsmath,amsfonts}
\usepackage{algorithmic}
\usepackage{algorithm}
\usepackage{array}
\usepackage[caption=false,font=normalsize,labelfont=sf,textfont=sf]{subfig}
\usepackage{textcomp}
\usepackage{stfloats}
\usepackage{url}
\usepackage{verbatim}
\usepackage{graphicx}
\usepackage{cite}
\usepackage{color}
\usepackage{multirow}
\usepackage{bbding}
\usepackage{fontawesome}
\usepackage{utfsym}
\usepackage[switch]{lineno}
\usepackage{tabularx}
\begin{document}
\newcommand{\etal}{\emph{et al.}}

\title{Adaptive Surface Normal Constraint for Geometric Estimation from Monocular Images}

\author{Xiaoxiao Long$^{\ast}$, Yuhang Zheng$^{\ast}$, Yupeng Zheng, Beiwen Tian, Cheng Lin, \\
Lingjie Liu,
Hao Zhao$^{\dagger}$, Guyue Zhou, Wenping Wang$^{\dagger}$,~\IEEEmembership{Fellow,~IEEE} 
\IEEEcompsocitemizethanks{
\IEEEcompsocthanksitem Xiaoxiao Long and Yuhang Zheng contribute equally to this work. Xiaoxiao Long is with the Department of Computer Science, the University of Hong Kong. Yuhang Zheng is with the Institute for AI Industry Research, Tsinghua University. 

\IEEEcompsocthanksitem Hao Zhao and Wenping Wang are the corresponding authors.\protect\\ Email: zhaohao@air.tsinghua.edu.cn, wenping@cs.hku.hk

\IEEEcompsocthanksitem Yupeng Zheng, Beiwen Tian, Hao Zhao, and Guyue Zhou are with the Institute for AI Industry Research, Tsinghua University. 

\IEEEcompsocthanksitem Cheng Lin is with Tencent Games.

\IEEEcompsocthanksitem Lingjie Liu is with the Department of Computer and Information Science, the University of Pennsylvania.

\IEEEcompsocthanksitem Wenping Wang is with the Department of Computer Science \& Engineering, Texas A\&M University. 
}}

\markboth{IEEE TRANSACTIONS ON PATTERN ANALYSIS AND MACHINE INTELLIGENCE}%
{Shell \MakeLowercase{\textit{et al.}}: A Sample Article Using IEEEtran.cls for IEEE Journals}


\IEEEtitleabstractindextext{%
\begin{abstract}
We introduce a novel approach to learn geometries such as depth and surface normal from images while incorporating geometric context. The difficulty of reliably capturing geometric context in existing methods impedes their ability to accurately enforce the consistency between the different geometric properties, thereby leading to a bottleneck of geometric estimation quality. We therefore propose the Adaptive Surface Normal (ASN) constraint, a simple yet efficient method. Our approach extracts geometric context that encodes the geometric variations present in the input image and correlates depth estimation with geometric constraints. By dynamically determining reliable local geometry from randomly sampled candidates, we establish a surface normal constraint, where the validity of these candidates is evaluated using the geometric context. Furthermore, our normal estimation leverages the geometric context to prioritize regions that exhibit significant geometric variations, which makes the predicted normals accurately capture intricate and detailed geometric information. Through the integration of geometric context, our method unifies depth and surface normal estimations within a cohesive framework, which enables the generation of high-quality 3D geometry from images. We validate the superiority of our approach over state-of-the-art methods through extensive evaluations and comparisons on diverse indoor and outdoor datasets, showcasing its efficiency and robustness.
Code and data are available at \url{https://www.xxlong.site/ASNDepth/}.

\end{abstract}
}
\maketitle
\IEEEdisplaynontitleabstractindextext
\IEEEpeerreviewmaketitle
\section{Introduction}


\IEEEPARstart{E}{stimating} geometric information such as depth and surface normal from monocular color images is a fundamental task in computer vision, which benefits lots of downstream applications such as structure from motion (SfM)~\cite{liu2022depth, torresani2008nonrigid}, 3D reconstruction~\cite{choe2021volumefusion, slavcheva2020variational}, pose estimation~\cite{liu2022gen6d}, simultaneous localization and mapping (SLAM)~\cite{teed2021droid, tateno2017cnn}, and autonomous driving~\cite{zheng2023steps, menze2015object}. 
This task is challenging because estimating geometry from a single image is an ill-posed problem that requires extensive knowledge of the 3D world.

\begin{figure}[t]
\setlength{\abovecaptionskip}{0pt}
\setlength{\belowcaptionskip}{0pt}
    \centering
    \includegraphics[width=\columnwidth]{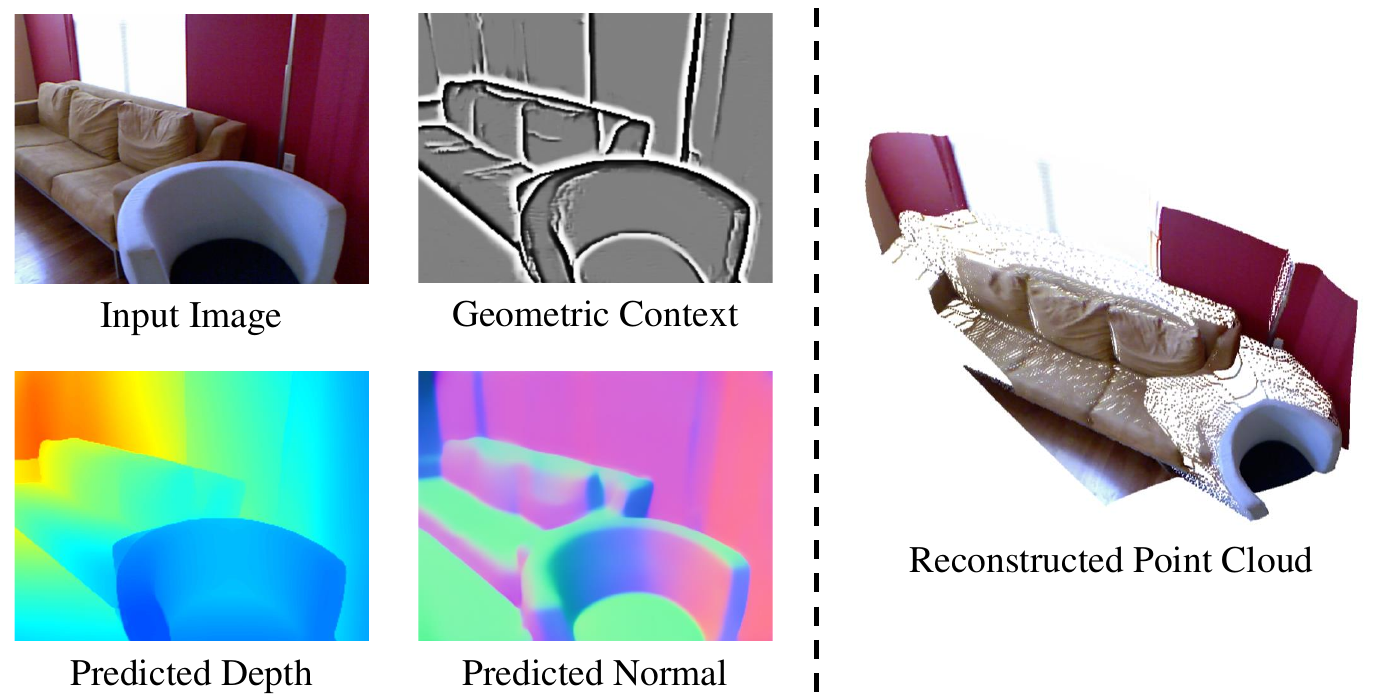}
    \caption{Taking a monocular RGB image as input, our approach first produces a geometric context map that encodes 3D geometric variances, and then jointly predicts depth and surface normal in a geometry-aware manner. Specifically, we rely on the geometric context to enforce Adaptive Surface Normal constraint on the predicted depth, which enables the predicted depth to faithfully preserve 3D geometry, thus yielding a high-quality point cloud converted from the depth. Meanwhile, we perform surface normal estimation guided by the geometric context, and the predicted normal map can keep rich geometric details.}
    \label{fig:teaser}
    \vspace{-3mm}
\end{figure}

Recent advancements in deep learning have significantly contributed to the field of estimating geometry from monocular images. While numerous methods~\cite{eigen2014depth, liu2015learning, fu2018deep, roy2016monocular, ranftl2016dense, wang2015designing, ladicky2014discriminatively} have demonstrated compelling results in image-based geometry estimation, most of these approaches are designed to address either depth estimation or normal estimation separately, overlooking the geometric correlation between the two tasks. Additionally, these methods often focus on minimizing pixel-wise errors in 2D image space, which therefore cannot faithfully capture the geometric features defined in 3D space.

The two geometric representations, depth and normal, are indeed correlated. First, surface normals can be estimated by the 3D points obtained from depth information. Second, surface normals are determined by a surface tangent plane, which inherently encodes local geometric context~\cite{hoiem2005geometric}, such as the derivatives of depth.
The work Taskonomy~\cite{zamir2018taskonomy} has demonstrated that joint learning of correlated tasks, like depth and normal estimation, can enhance the performance of each task compared to learning them separately. This joint learning approach leverages the ability of the two tasks to propagate information and provide constraints to each other through gradient backpropagation, thereby boosting overall performance.

Some latest efforts propose to incorporate geometric constraints into the joint learning of depth and normal, where surface normal is extracted from depth maps as constraints to regularize depth for capturing 3D geometric features. 
Various strategies has been proposed to extract surface normal, including random sampling~\cite{yin2019enforcing}, Sobel-like operator~\cite{hu2019revisiting, kusupati2020normal} and differentiable least squares~\cite{qi2018geonet, long2020occlusion}. 
However, one critical issue remains unsolved, i.e., how to determine the reliable local geometry, to correlate surface normal with depth. For example, at shape boundaries or corners, the neighboring pixels for a depth map belong to different geometries, where the local plane assumption is not satisfied. Due to this challenge, these methods either struggle to capture the local features~\cite{yin2019enforcing}, or are sensitive to local geometric variations (noises or boundaries)~\cite{hu2019revisiting, kusupati2020normal}. 
While some prior works, spanning optical flow~\cite{revaud2015epicflow, xiao2006bilateral, ren2008local, bao2014fast} to stereo matching~\cite{jiao2014local, qi2020geonet++}, have attempted to incorporate shape-aware regularization into their pipelines, they only focus on 2D feature edges, such as Canny edges, derived from image intensity variation, without considering the geometric structures of shapes in 3D space.

In this paper, we introduce a unified scheme to jointly predict depth and surface normal from single image, specifically a novel adaptive surface normal constraint is proposed to derive geometric context from images that correlates depth estimation and normal estimation together (as illustrated in Fig.~\ref{fig:teaser}).
The geometric context encodes local 3D geometric variations, and can provide informative hints for geometric estimation.

Our key idea is to adaptively determine the faithful local geometry indicated by the learned geometric context to correlate depth and surface normal.
Specifically, 1) for a target point on the predicted depth map, we randomly sample a set of point triplets in its neighborhood to define the candidates of normals. 2) We determine the confidence score of each normal candidate by measuring the similarity of learned latent feature, i.e., geometric context, between the candidate and the target point. The normal is adaptively estimated as a weighted sum of all the candidates. 3) We enforce the consistency between the recovered normal and ground truth surface normal. 
In order to produce accurate recovered normal, the geometric context will be automatically trained to capture the geometric variations of the input image.
As a result, our method can not only accurately capture sufficient geometric context information but also be highly efficient for computation. Furthermore, the weighted effect of the confidence scores makes the recovered normal more robust to local variations, e.g., noises, boundaries and sharp changes.

The learned geometric context is utilized to improve the predicted normals by incorporating rich geometric details. This context effectively highlights regions with high-frequency geometric variations, such as boundaries and small structures, where surface normals often exhibit discontinuities.
Given that regions with rich details constitute only a small portion of the entire image, neural networks may struggle to accurately predict surface normals in these areas. To address this, we propose a geometric context-guided approach to enhance the predicted normals with geometric details. Specifically, leveraging the geometric context, we sample pixels with rich details from the input image. Subsequently, we perform a two-stage estimation on these pixels to obtain their normals. Finally, we assign larger loss weights to the sampled pixels, prompting the neural network to focus on these regions during training.


Our simple strategy has some unique advantages for geometric estimation: 
\textbf{1)} we measure the confidence using the learned geometric context, of which representational capacity is applicable to complex structures and informative to correlate the surface normal constraint with the predicted depth, which makes predicted depth shape-aware.  \textbf{2)} With the guidance of learned geometric context, the predicted surface normal has significantly improvement, including being more robust to high-frequency geometric variations and more accurate in regions with rich geometric details.
\textbf{3)}, our method achieves superior results on the public indoor and outdoor datasets and considerably outperforms the state-of-the-art methods.


\vspace{1.5mm}
\noindent \textbf{Difference from our Conference Paper:~} This paper signiﬁcantly improves the conference version\cite{long2021adaptive}: 
i) a unified scheme to jointly estimate depth and surface normal in the guidance of geometric context;
(ii) we propose a context guided normal estimation network to improve the normal prediction for capturing rich geometric details; (iii) we prove that the ASN module is effective on both transformer backbone and CNN backbone; 
(iv) from both qualitative and quantitative perspectives, our results are signiﬁcantly better than our conference version ASN~\cite{long2021adaptive}, especially on 3D metrics; 
(v) we show that our method not only works well in indoor scenes but also proves robustness in outdoor scenes.
The rest of the paper is organized as follows. Sec.~\ref{related_work} reviews the related work on depth and surface normal prediction. In Sec.~\ref{method}, we elaborate the details of the Adaptive Surface Normal constraint. In Sec.~\ref{setup}, we introduce experimental datasets and evaluation metrics. The results of our methods are shown in Sec.~\ref{exp} and more detailed analysis are presented in Sec.~\ref{discuss}. Finally, in Sec.~\ref{clu}, we draw the conclusions.

\section{Related Work}
\label{related_work}
\begin{figure}[t]
\setlength{\abovecaptionskip}{0pt}
\setlength{\belowcaptionskip}{0pt}
    \centering
    \includegraphics[width=0.9\linewidth]{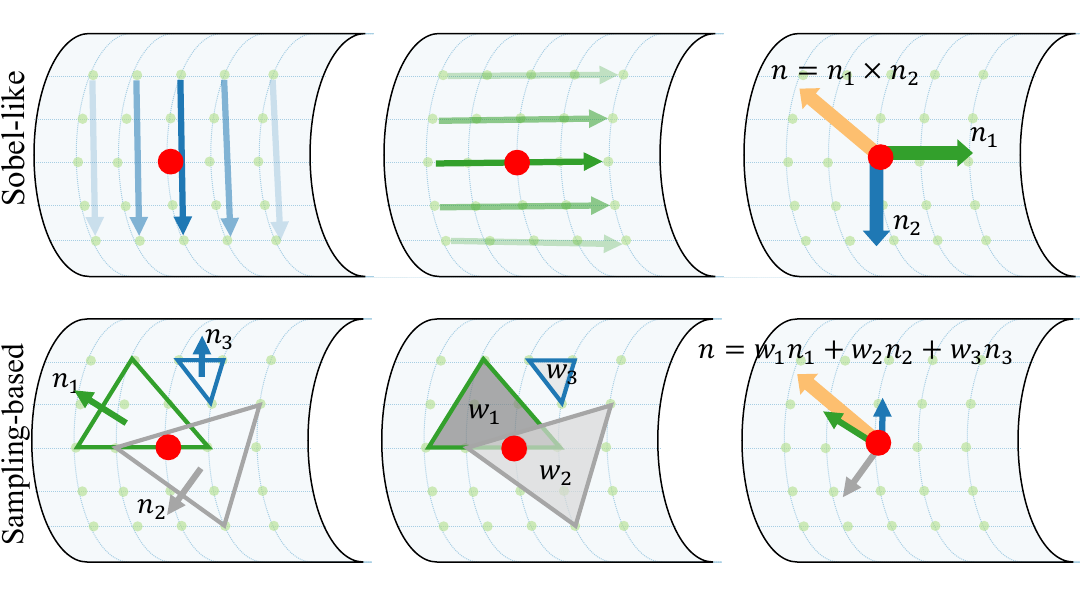}
    \caption{Sobel-like operator versus ours for surface normal calculation. The Sobel-like operator first calculates two principle vectors along up-down and left-right directions, and then use their cross product to estimate the normal. Ours first computes the normal vectors of the randomly sampled triplets, and then adaptively combines them together to obtain the final result.}
    \label{fig:normal_diagram}
    \vspace{-3mm}
\end{figure}

\subsection{Monocular depth estimation} As an ill-posed problem, monocular depth estimation is challenging, given that minimal geometric information can be extracted from a single image. Recently, benefiting from the prior structural information learned by the neural network, many learning-based works~\cite{eigen2014depth,liu2015learning,xu2017multi,fu2018deep,roy2016monocular,ranftl2016dense,godard2019digging,godard2017unsupervised,liu2021fcfr,liu2021learning} have achieved promising results.
Eigen~\etal~\cite{eigen2014depth} directly estimate depth maps by feeding images into a multi-scale neural network.
Laina~\etal~\cite{laina2016deeper} propose a deeper residual network and further improve the accuracy of depth estimation. Liu~\etal~\cite{liu2015learning} utilize a continuous conditional random field (CRF) to smooth super-pixel depth estimation. Xu~\etal~\cite{xu2017multi} propose a sequential network based on multi-scale CRFs to estimate depth. Fu~\etal~\cite{fu2018deep} design a novel ordinal loss function to recover the ordinal information from a single image. Unfortunately, the predicted depth maps of these methods always fail to recover important 3D geometric features when converted to point clouds, since these methods do not consider any geometric constraints.

\subsection{Joint depth and normal estimation} Since the depth and surface normal are closely related in terms of 3D geometry, there has been growing interests in joint depth and normal estimation using neural networks to improve the performance. Several works \cite{eigen2015predicting,zhang2019pattern,xu2018pad,li2015depth} jointly estimate depth and surface normal using multiple branches and propagate the latent features of each other. Nevertheless, since there are no explicit geometric constraints enforced on the depth estimation, the predicted geometry of these methods is still barely satisfactory. 

Consequently, methods~\cite{xu2019depth,yang2018unsupervised,qiu2019deeplidar,hu2019revisiting,qi2018geonet,long2020occlusion,kusupati2020normal} are proposed to explicitly enforce geometric constraints on predicted depth maps. Hu~\etal~\cite{hu2019revisiting} and Kusupati~\etal~\cite{kusupati2020normal} utilize a Sobel-like operator to calculate surface normals from predicted depth, and then enforce them to be consistent with the ground truth. Nonetheless, the Sobel-like operator can be considered as a fixed filter kernel that indiscriminately acts on the whole image (see Fig.~\ref{fig:normal_diagram}), leading to unacceptable inaccuracy and sensitivity to noises. To constrain surface normal more reliably, Qi~\etal~\cite{qi2018geonet} and Long~\etal~\cite{long2020occlusion} propose to utilize a differentiable least square module for surface normal estimation. These methods optimize the geometric consistency, of which solution is more accurate and robust to noises but limited to expensive computation. Yin~\etal~\cite{yin2019enforcing} introduce virtual normal, which is a global geometric constraints derived from the randomly sampled point triplets from predicted depth. However, this constraint struggles to capture local geometric features, given that the point triplets are randomly sampled from the whole image.

\subsection{Edge preserving methods} Out of the statistical relations between shape boundaries and image intensity edges, many works leverage this statistic prior to benefit many vision tasks.
Anisotropic diffusion~\cite{perona1990scale,black1998robust,weickert1998anisotropic} is a well-known technique for image denoising without removing important details of image content, typically edges.
Works~\cite{revaud2015epicflow,xiao2006bilateral,ren2008local,bao2014fast} propose variational approaches with anisotropic diffusion to model local edge structures for optical flow estimation. Some stereo/monocular depth estimation works rely on pretrained edge detection network~\cite{song2018edgestereo} or Canny edge detector~\cite{jiao2014local,qi2020geonet++}, to extract image edges to improve depth estimation. 
However, only a small fraction of the intensity edges keep consistent with true geometric shape boundaries. Our method could detect the true shape boundaries where 3D geometry changes instead of image intensity edges. Our recent method DPF~\cite{chen2023dpf} exploits 2D coordinate-based (or say implicit) networks to encourage edge-preserving smoothness.

\section{Method}
\label{method}
\begin{figure*}[t]
\setlength{\abovecaptionskip}{0pt}
\setlength{\belowcaptionskip}{0pt}
    \centering
    \includegraphics[width=\linewidth]{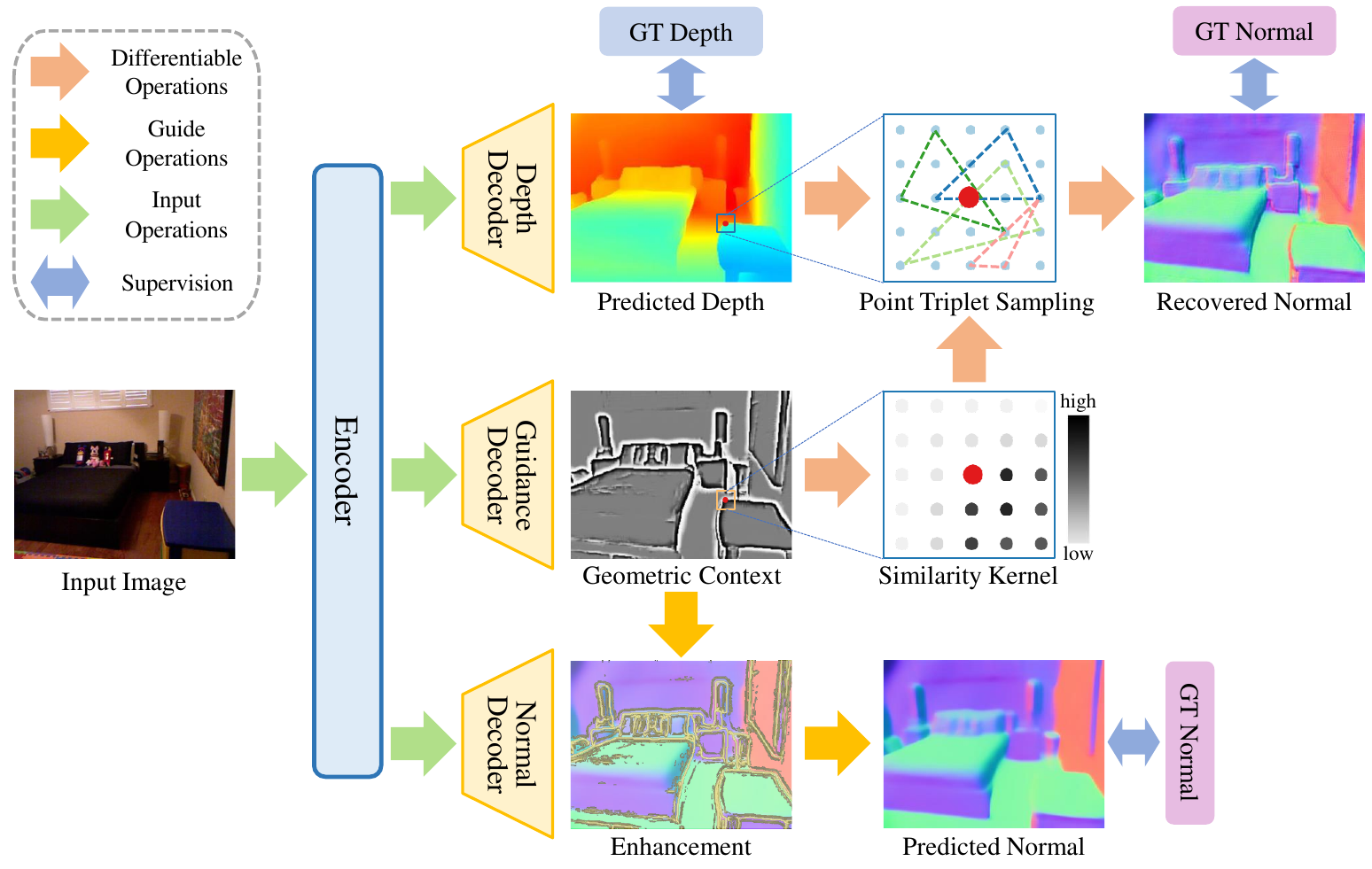}
    \caption{Overview of our method. Taking a single image as input, our model produces depth maps, geometric context, and surface normal maps from three decoders, respectively. We recover surface normal from the predicted depth map with our proposed Adaptive Surface Normal (ASN) computation method. The similarity kernels computed from geometric context enable our surface normal calculation to be aware of local geometry, like shape boundaries and corners. Furthermore, geometric context encodes the rich geometric variances which the predicted surface normal usually struggles to capture. Thus, we design an approach, using the geometric context to guide the surface normal estimation. Finally, pixel-wise depth/normal supervision is enforced on the predicted depth/normal, while the geometric supervision is enforced on the recovered surface normal.}
    \label{fig:pipeline}
    \vspace{-3mm}
\end{figure*}

The overall pipeline is illustrated in Fig.~\ref{fig:pipeline}. Given a single color image $I$ as input, our method jointly estimates its depth map $D_{\mathrm{pred}}$ and normal map $N_{\mathrm{pred}}$, where the two tasks are correlated together to further enhance their 3D geometric properties with the guidance of learned geometric context map $F_{\mathrm{geo}}$.
Besides, we recover the surface normal from the predicted depth using the proposed \textit{adaptive strategy} for constraint, which we refer to as Adaptive Surface Normal (ASN) constraint. To avoid misunderstandings, we use the predicted surface normal (denoted as $N_{\mathrm{pred}}$) to represent the normal maps directly produced by neural networks as predicted surface normal and use the recovered surface normal (denoted as $N_{\mathrm{recover}}$) to represent the surface normal calculated from the depth.

For depth estimation, first, like most works in this field, we employ pixel-wise depth supervision like $SILog$ loss over the predicted depth $D_{\mathrm{pred}}$ and ground truth depth $D_{\mathrm{gt}}$. 
Moreover, we compute the recovered surface normal $N_{\mathrm{recover}}$ from $D_{\mathrm{pred}}$ using the proposed ASN method and enforce consistency between $N_{\mathrm{recover}}$ and the ground truth surface normal $N_{\mathrm{gt}}$.
For normal estimation, we propose a novel approach, which leverages the guidance of geometric context to further enhance the capability of capturing geometric details.



\subsection{Adaptive Surface Normal constraint}

\subsubsection{Local plane assumption}
To correlate surface normal constraint with depth estimation, we adopt the local plane assumption following ~\cite{qi2018geonet,long2020occlusion}. That is, a small set of neighborhoods of a point forms a local plane, of which normal vector approximates the surface normal. Hence, for a pixel on the depth map, its surface normal can be estimated by the local patch formed by its neighboring points. In theory, the local patch could have arbitrary shapes and sizes. In practice, however, square local patches are widely adopted with sizes $(2m+1)\times(2m+1), m=1,2,...,n$, due to its simplicity and efficiency.

\subsubsection{Normal candidates sampling}\label{subsect:full-model}
To compute the surface normal, unlike prior works that typically utilize least square fitting ~\cite{qi2018geonet,long2020occlusion} or Sobel-like kernel approximation ~\cite{hu2019revisiting,kusupati2020normal}, we propose a novel strategy based on random sampling. 

For a target point $P_i\in\mathbb{R}^3$, we first extract all the points $\mathbb{P}_i=\left\{P_{j} \mid P_j\in\mathbb{R}^3,\ j=0,\ldots, r^2-1\right\}$ within a local patch of size $r \times r$. Then, we randomly sample $K$ point triplets in $\mathbb{P}_i$. All sampled point triplets for the target point $P_i$ are denoted as $\mathbb{T}_i=\left\{\left(P_k^A, P_k^B, P_k^C\right) \mid P\in \mathbb{R}^3,\ k=0,\ldots ,K-1\right\}$.
If the three points are not colinear, the normal vector of the sampled local plane can be directly computed by the cross-product:
\begin{equation}
\label{cross_normal}
\vec{n}_{k}=\frac{\overrightarrow{P_k^A P_k^B} \times \overrightarrow{P_k^A P_k^C}}{\mid \overrightarrow{P_k^A P_k^B} \times \overrightarrow{P_k^A P_k^C} \mid}.
\end{equation}

A normal vector will be flipped according to the viewing direction if it does not match the camera orientation. In this way, for each target point, we obtain $K$ normal candidates corresponding to $K$ sampled local planes. Next, we adaptively determine the confidence of each candidate to derive the final normal estimation result.

\subsubsection{Geometric context adaption}
We observe that the neighbors of a target point may not lie in the same tangent plane, especially at a region where the geometry changes, e.g., shape boundaries or sharp corners. Thus, we propose to learn a geometric context map that is context aware to reflect the geometric variations. Therefore, the network can determine the confidence of the neighboring geometry by measuring the learned  geometric context.

As shown in Figure~\ref{fig:pipeline}, the geometric context is predicted by a guidance decoder, where the geometric context is a latent feature map with size $H\times W \times C$. Initially, the predicted latent feature map will not contain any informative contents that encode the geometric variations if without any supervisions. 
However, there are no ground truth geometric context to supervise the training, and therefore we propose the adaptive surface normal constraint that could encourage the geometric content feature map to encode the geometric information by just enforcing the consistency between the recovered normal from depth and the ground truth normal. 
We discuss how the strategy works in the following paragraphs.

Given the learned geometric context map, we measure the $L_2$ distance of the features of a sampled point $P_{j}$ and the target point $P_{i}$, and then use a normalized Gaussian kernel function to encode their latent distance into $[0,1]$:
\begin{equation}
\label{kernel_function}
\begin{aligned}
\mathcal{L}\left(P_{i}, P_{j}\right) &= e^{-0.5\left\|F_{\mathrm{geo}}\left(P_{i}\right)-F_{\mathrm{geo}}\left(P_{j}\right)\right\|_{2}} \\
\overline{\mathcal{L}}\left(P_{i}, P_{j}\right) &= \frac{\mathcal{L}\left(P_{i}, P_{j}\right)}{\sum_{P_n \in \mathbb{P}_i} \mathcal{L}\left(P_{i}, P_{n}\right)},
\end{aligned}
\end{equation}
where $F_{\mathrm{geo}}(\cdot)$ is the geometric context map, $\left\| \cdot \right\|_{2}$ is $L_2$ distance, and $\mathbb{P}_i$ is the neighboring point set in the local patch of $P_i$ as aforementioned. E.q.~\ref{kernel_function} gives a confidence score, where the higher the confidence is, the more likely the two are to locate in the same tangent plane with the target pixel. Accordingly, the confidence score of a local plane $(P_k^A, P_k^B, P_k^C)$ to the center point $P_i$ given by the geometric adaption is defined by:

\begin{equation}
\label{geo_adaption}
g_k = \prod_{t=A,B,C} \overline{\mathcal{L}}\left(P_{i}, P_{k}^t\right).
\end{equation}

This is the multiplication of three independent probabilistic scores of the three sampled points, which measures the reliability of a sampled local plane.

\subsubsection{Area adaption}
The area of a sampled local plane (triangle) is an important reference to determine the reliability of the candidate. A larger triangle captures more information and thus would be more robust to local noise, as shown in ~\cite{yin2019enforcing}. For a triangle $T_k$, we simply consider its projected area $s_k$ on the image as a measurement of the confidence score. Note that the area is calculated on the 2D image, since the sampled triangles in the 3D space could be very large due to depth variations, leading to unreasonable overestimation. 

\noindent 
Finally, the normal for a point $P_i$ is determined by a weighted combination of all the $K$ sampled candidates, where the weights represent the confidence given by our adaptive strategy:
\begin{equation}
\label{full_adaption}
\vec{n}_i=\frac{\sum_{k=0}^{K-1} s_{k} \cdot g_k \cdot \vec{n}_{k}}{\sum_{k=0}^{K-1} s_{k} \cdot g_k},
\end{equation}
where $K$ is the number of sampled triplets, $s_k$ is the projected area of three sampled point $(P_k^{A}, P_k^{B}, P_k^C)$ on the 2D image, and $n_k$ is its normal vector.

\begin{figure}[!t]
\setlength{\abovecaptionskip}{0pt}
\setlength{\belowcaptionskip}{0pt}
    \centering
    \includegraphics[width=\linewidth]{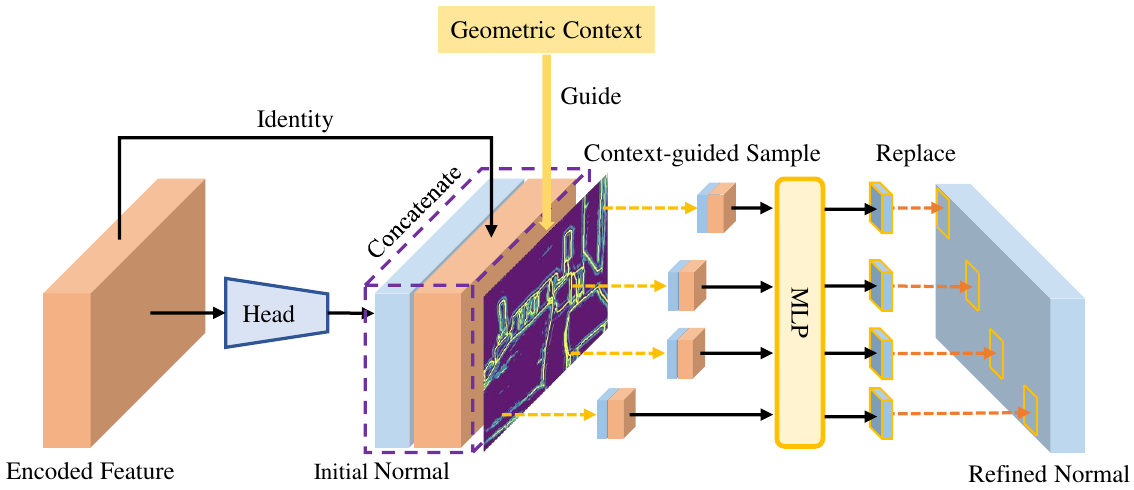}
    \caption{The structure of our proposed normal-estimation approach. Taking the feature maps produced by the encoder as input, the module first generates initial surface normal predictions using a normal head and then concatenates the encoder feature maps with the surface normals. Next, we use geometric context to guide the sampling of pixels that locates in regions with rich geometric details. The sampled pixels are taken into an MLP to output pixel-wise refined surface normals. Finally, the initial normals of the sampled pixels are replaced with the newly refined normals.
    }
    \label{fig:normal_decoder}
    \vspace{-8mm}
\end{figure}

With the recovered normal obtained from the estimated depth, we enforce the consistency between the recovered normal and the ground truth normal.
If simply holding the local plane assumption, the recovered normal always contain inaccurate approximation around geometric boundaries. 
To minimize the normal errors, the latent feature map will be encouraged to be aware of the local planes, where the features of pixels in the same local plane share high similarity but those in different local planes present contrast as defined in Eq.~\ref{kernel_function}. As a result, the proposed surface normal constraint can adaptively select pixels sharing high similarity that infer they are in the same local plane to approximate surface normal.

\vspace{-2mm}
\subsection{Geometric context guided normal estimation}
\subsubsection{The structure of normal estimation branch}
Most prior works typically treat every pixel equally when calculating surface normal loss, which leads to inferior performance in some challenging regions with geometric details such as geometric boundaries and intricate structures. This is because the images predominantly contain man-made scenes that are dominated by planes where surface normals have few variations but play a leading role in the training. As a result, the network leans to ignore high-frequency geometric features, which shows poor ability to capture detailed geometries. Fortunately, the learned geometric context, which indicates geometric details, can be reused to guide the network to focus on challenging regions. Based on this, we propose a geometric context guided approach for surface normal estimation to tackle this problem. In simple terms, the loss weights assigned to each pixel are determined by its corresponding value in geometric context map. Thus, the impact of geometric details improves during training.

The detailed structure of our proposed method is illustrated in Fig. ~\ref{fig:normal_decoder} where the process of normal estimation is composed of four steps: \textbf{1)} take encoder features as input and directly output initial surface normals; \textbf{2)} concatenate the initial normals and the encoder features to form a residual block; \textbf{3)} sample pixels in the regions which contain rich geometric details with the guidance of geometric context map; \textbf{4)} an MLP takes the concatenated features of the sampled pixels to output refined normals, and then replace the initial predictions of the sampled pixels by the newly refined normals. We will elaborate on how to perform the sampling below.

\subsubsection{Context-guided sampling}
\label{context_guided_sampling}
Our learned geometric context map encodes the geometric variations of the captured scene. However, the context map has multiple channels and doesn't explicitly indicate the locations of the geometric details.
Therefore, we first perform absolute summation on the context map along its channel dimension to get an intensity map, and then calculate the first-order or second-order derivatives of the intensity map in both horizontal and vertical direction, finally yielding a gradient magnitude map or a curvature magnitude map. The magnitude map reflects the confidence of a pixel being in the regions with geometric details, which enables to leverage the magnitudes as guidance weights to perform pixels sampling.
For the sampling operation, we sort all the pixels of an image by their corresponding weights, and sample the top $v$ pixels. 


\vspace{-2mm}
\subsection{Network architecture} Our network adopts a multi-task structure, which consists of one encoder and three decoders. We use our prior work Cerberus~\cite{chen2022cerberus}, which is inherited from Dense Prediction Transformer (DPT)~\cite{ranftl2021vision}, as the backbone. Taking one image as input, the transformer produces feature maps for four scales with resolutions of 1/16, 1/8, 1/4, and 1/2. We then adpot three decoders to estimate depth and surface normal and capture geometric context, respectively. 

\begin{figure*}[!t]
\setlength{\abovecaptionskip}{1pt}
\setlength{\belowcaptionskip}{1pt}
    \centering
    \includegraphics[width=\linewidth]{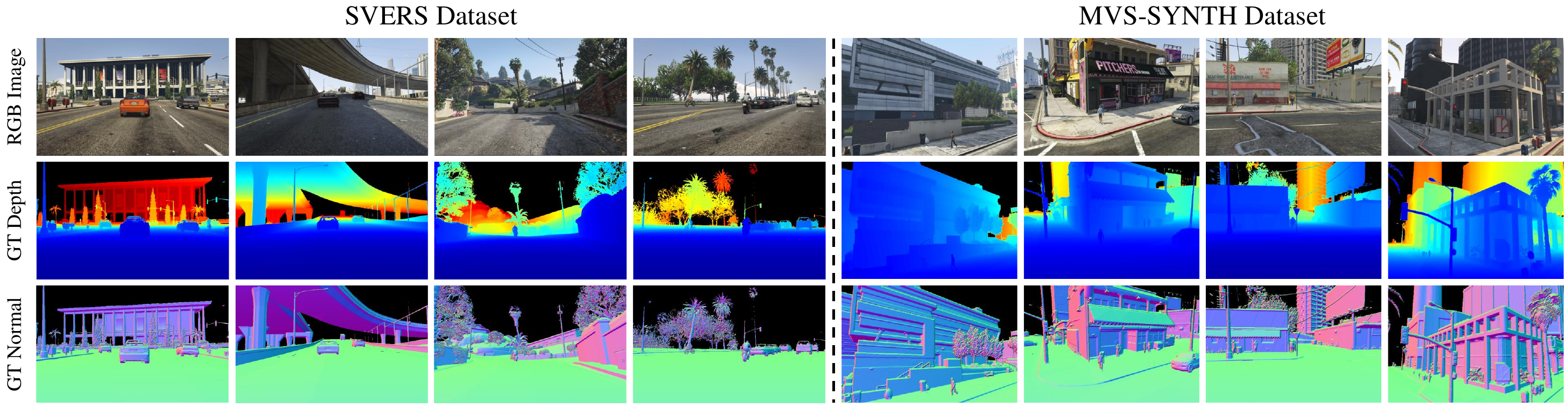}
    \caption{Some samples of SVERS dataset and MVS-SYNYH dataset. 
    We mask the regions whose depth values are larger than 80m as black for better visualization.
    Although both datasets are generated from the GTA-V game, SVERS contains diverse driving scenarios with vehicles, motorcycles, and pedestrians in urban, overpasses, and countryside environments, while MVS-SYNTH mainly consists of buildings, vehicles, and pedestrians in urban areas with varying viewpoints.}
    \label{fig:dataset}
    \vspace{-3mm}
\end{figure*}

\subsubsection{Depth decoder}
The depth decoder produces coarse-to-fine predicted depth in four scales. It consists of four blocks in different scales, each of which is constituted by two ResNet~\cite{he2016deep} basic blocks with a channel-reducing convolution layer. The appending convolution layers are used to produce one-channel output for depth estimation. 

\subsubsection{Guidance decoder}
The guidance decoder is used to generate the geometric context that capture geometric context. The guidance decoder adopts a nearly identical structure with the depth decoder but without any feature-reducing layers. 

\subsubsection{Normal decoder}
The normal decoder produces coarse-to-fine predicted normals in four scales with resolutions of 1/8, 1/4, 1/2, and full size. Our proposed geometric context guided approach is repeatedly applied on the last three scales with relatively high resolutions. The normal estimation network employs a multilayer perceptron (MLP) with three hidden layers, and each layer consists of 128 channels with a ReLU activation function.

\vspace{-2mm}
\subsection{Loss functions} Our training loss has three types of terms: depth loss term, ASN loss term and surface normal loss terms. 
For the depth term, we apply the \textit{scale invariant logarithmic error} $(SILog)$ loss to the predicted depth maps of the four scales:

\begin{equation}
\label{depth_term}
    \mathcal{L}_{d} = \sum_{s=0}^{3} \lambda_{d}^{s-3} \left[\frac{1}{m_{s}}\sum_{i=1}^{m_{s}} e_{s,i}^{2} + \frac{1}{m_{s}^{2}}\left(\sum_{i=1}^{m_{s}} e_{s,i}\right)^{2}\right],
 \end{equation}
where $e_{s,i}=\mathrm{log}\left(\tilde{d}_{s,i}\right)-\mathrm{log}\left(d_{i}\right)$, $\tilde{d}_{s,i}$ means the predicted depth of pixel $i$ at $s^\mathrm{th}$ scale, $d_{i}$ is the ground truth depth of pixel $i$, $m_{s}$ denotes the number of pixels having valid
ground truth values at $s^\mathrm{th}$ scale, and $\lambda_{d}$ is a factor for balancing different scales in depth estimation. Here we set $\lambda_{d} = 0.8$.

To enforce geometric constraint on the predicted depth map, we compute the surface normals from the predicted depth map of the last scale with our proposed Adaptive Surface Normal calculation method. To minimize the differences between the computed surface normals with the ground truth, we adopt a cosine similarity loss:

\begin{equation}
\label{normal_term}
    \mathcal{L}_{asn} = \frac{1}{m_{s}}\sum_{i=1}^{m_{s}} \left(1 - \hat{n_i} \cdot n_i \right), s=3
\end{equation}
where $\hat{n_i}$ is the recovered normal map of the $i^\mathrm{th}$ pixel (surface normal calculated from the finest predicted depth map), and $n_i$ is the corresponding ground truth surface normal. 

For the predicted surface normal loss term, we calculate the loss for the predictions of all scales: 
\begin{equation}
\label{normal_term}
    \mathcal{L}_{n} = \sum_{s=0}^{3} \frac{\lambda_{n}^{s-3}}{m_{s}}\sum_{i=1}^{m_{s}} \left(1+w_{s,i} \right) \left(1 - \tilde{n}_{s,i}\cdot n_{s,i}\right),
\end{equation}
where $\tilde{n}_{s,i}$ is the prediction surface normal of the $i^\mathrm{th}$ pixel at $s^\mathrm{th}$ scale, $w_{s,i}$ is the importance weight of the $i^\mathrm{th}$ pixel at $s^\mathrm{th}$ scale that is derived from the guidance weight map (see Section~\ref{context_guided_sampling}). $\lambda_{n}$ is set to 0.8, a factor for balancing the losses of different scales for normal estimation. 

Therefore, the overall loss is defined as:
\begin{equation}
\label{full_loss}
    \mathcal{L} = \mathcal{L}_{d} + \alpha \mathcal{L}_{asn} + \beta \mathcal{L}_{n},
\end{equation}
where $\alpha$ and $\beta$ are set to 5 in all experiments, which are trade-off parameters to make the three types of terms roughly of the same scale.



\begin{figure*}[!t]
\setlength{\abovecaptionskip}{1pt}
\setlength{\belowcaptionskip}{1pt}
    \centering
    \includegraphics[width=\linewidth]{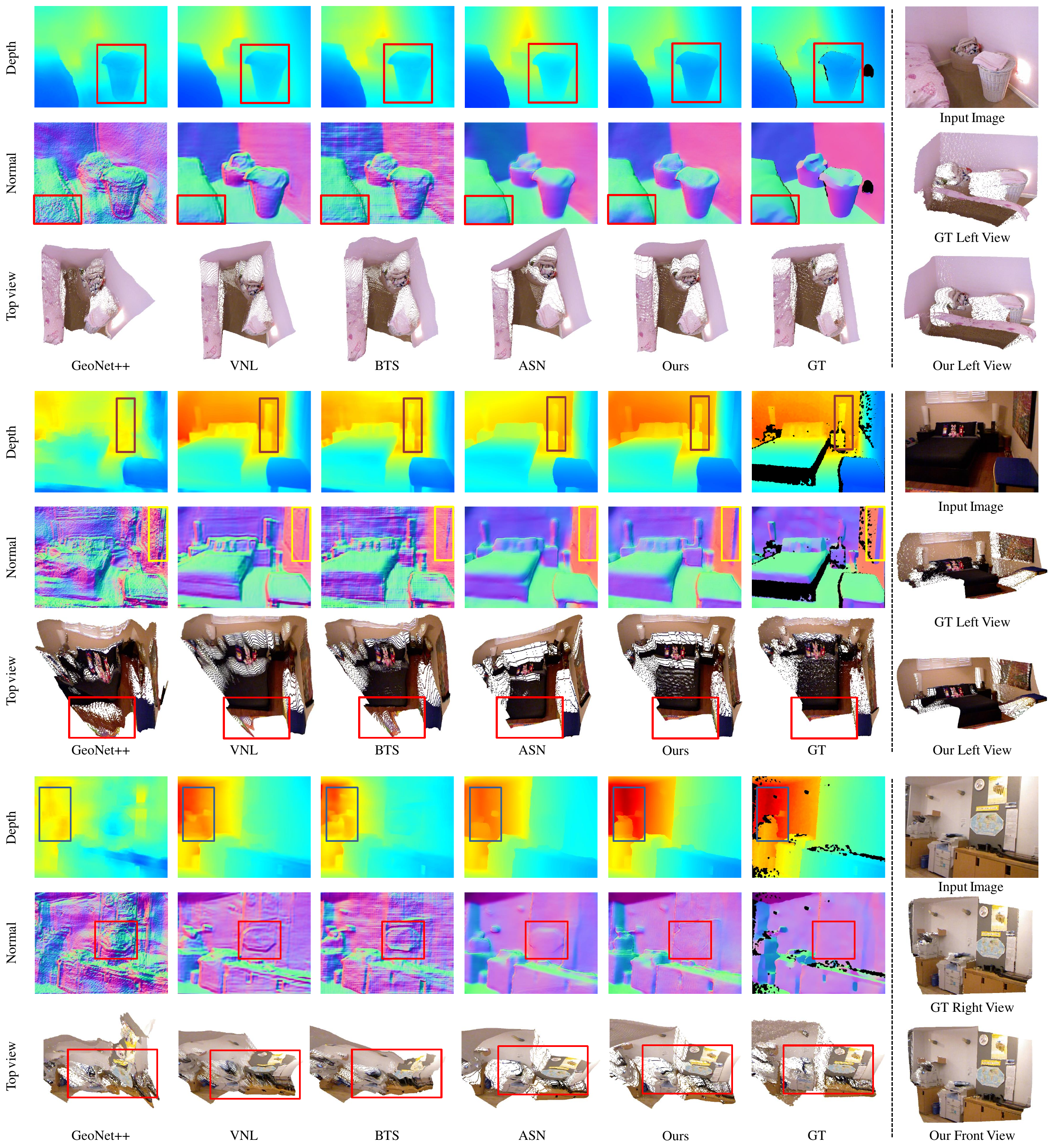}
    \caption{Qualitative comparisons with SOTAs on NYUD-V2. Compared with others, our predicted depth is more accurate and contains less noise. The recovered surface normal maps and point clouds demonstrate that our predicted depth faithfully preserves important geometric features. The black regions are invalid due to the lack of ground truth. The bounding boxes represent the regions where our method performs better than others.}
    \label{fig:nyu_compare}
    \vspace{-3mm}
\end{figure*}

\begin{figure*}[!t]
\setlength{\abovecaptionskip}{1pt}
\setlength{\belowcaptionskip}{1pt}
    \centering
    \includegraphics[width=\linewidth]{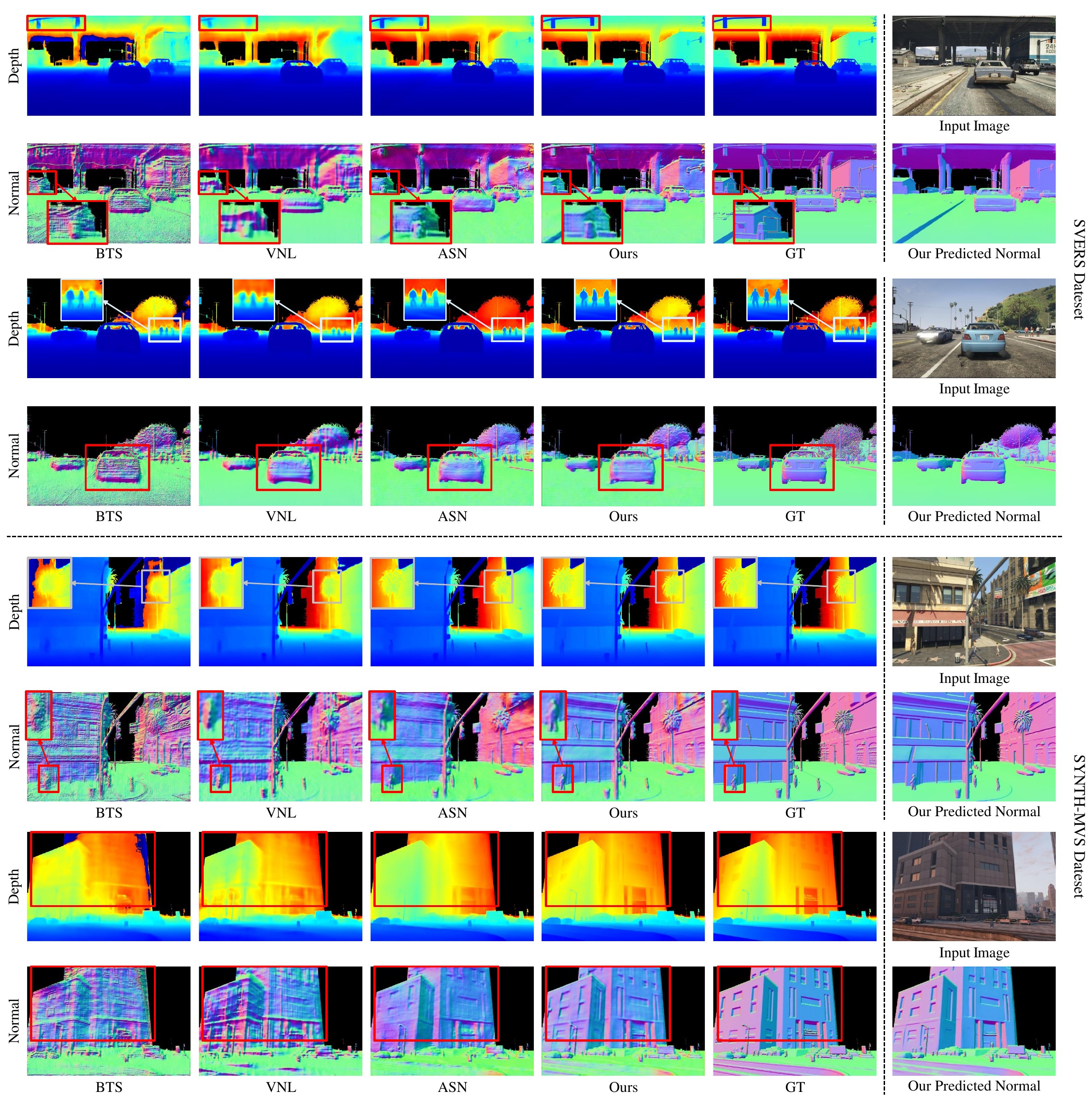}
    \caption{Qualitative comparisons with SOTAs on outdoor datasets. We show predicted depth and recovered surface normal for each method. 
    Compared with other methods, our predicted depth maps achieve higher accuracy. The recovered surface normal maps demonstrate that our predicted depth maps faithfully preserve important geometric features, such as small objects and thin structures. We further provide the predicted surface normal for reference. For better visualization, the regions whose depth values are beyond 80 meters are masked out as black. And the bounding boxes represent the regions where our method performs better than others. }
    \label{fig:SYNTH}
    \vspace{-3mm}
\end{figure*}

\begin{figure*}[!t]
\setlength{\abovecaptionskip}{1pt}
\setlength{\belowcaptionskip}{1pt}
    \centering
    \includegraphics[width=\linewidth]{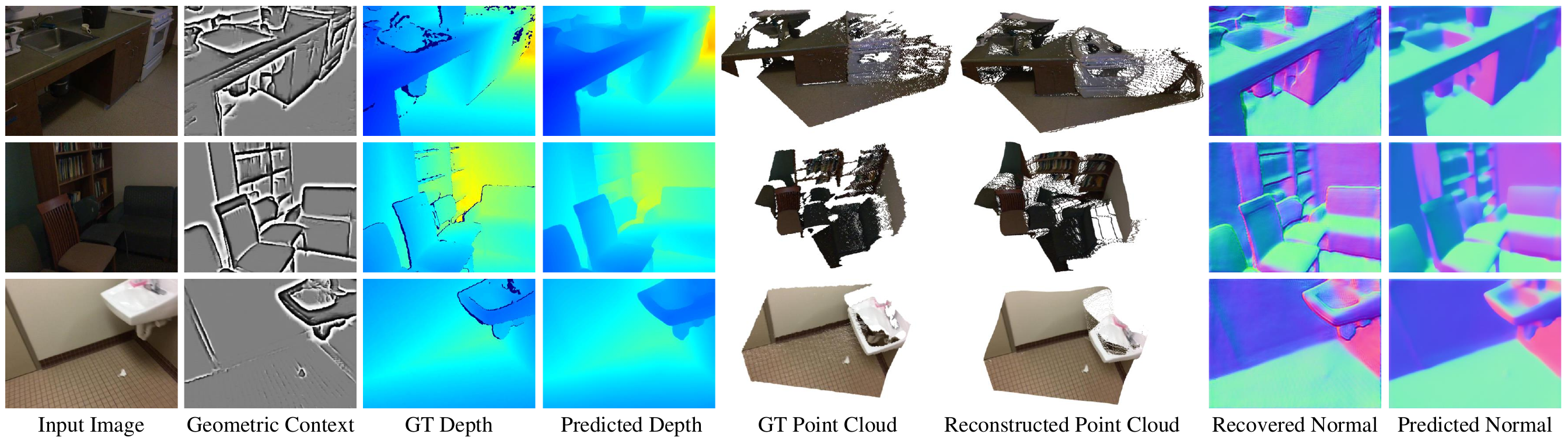}
    \caption{Qualitative results of our method on unseen ScanNet dataset. The second column showcases the effectiveness of the learned geometric context in accurately capturing geometric details, as described in Section~\ref{context_feature}. Additionally, we provide  visualizations of predicted depth, reconstructed point cloud, recovered normal, and predicted normal. Despite being applied to an unseen dataset, our method performs well in preserving geometric details.}
    \label{fig:scannet}
    \vspace{-3mm}
\end{figure*}

\vspace{-3mm}
\section{Experimental Setup}
\label{setup}

\subsection{Datasets}

\subsubsection{NYUD-V2} 
NYUD-V2 is a widely used indoor dataset and contains 464 scenes, of which 249 scenes are for training and 215 are for testing. We directly adopt the collected training data provided by Qi~\etal~\cite{qi2018geonet}, which has 30,816 frames sampled from the raw training scenes with precomputed ground truth surface normal. The precomputed surface normal is generated following the procedure of ~\cite{fouhey2013data}. Note that DORN~\cite{fu2018deep}, Eigen~\etal~\cite{eigen2015predicting}, Xu~\etal~\cite{xu2017multi}, Laina~\etal~\cite{laina2016deeper}, and Hu~\etal~\cite{hu2019revisiting} use 407k, 120k, 90k, 95k and 51k images for training, which are all significantly larger than ours. For testing, we utilize the official test set that is the same as the competitive methods, which contains 654 images.
\vspace{-2mm}
\subsubsection{ScanNet}  
In addition to the aforementioned evaluations, we further validate the performance of our method using the ScanNet dataset~\cite{dai2017scannet}. ScanNet is a widely used indoor scene dataset comprising over 1600 scenes. We specifically utilize the official test split of 100 scenes for evaluation purposes. From this test split, we uniformly select 2,167 images to conduct cross-dataset evaluations.
\vspace{-2mm}
\subsubsection{MVS-SYNTH}  
To further validate the effectiveness and robustness of our method, we evaluate our method on a more challenging outdoor dataset. Prior work shows that the noise of ground truth in real data leads to inferior performance while synthetic datasets can be used to solve such problem~\cite{hoiem2005geometric}. Thus, we choose MVS-SYNTH dateset~\cite{huang2018deepmvs} to evaluate our approach. It contains rich realistic urban scenes captured from varying camera viewpoints, showing high diversity and an acceptable domain gap with real data. MVS-SYNTH dataset consists of 120 sequences of urban scenes captured in the video game \textit{Grand Theft Auto V}~\cite{gtav}. Each sequence contains 100 RGB frames with a resolution of 810*540, ground truth depth maps, and intrinsic camera parameters. We then calculate surface normal maps from the GT depth maps as the GT normal. In our experiments, we randomly choose 6,000 frames for training and 960 frames for testing. The frames are resized to 672*448 pixels for memory-efficient training and testing.


\vspace{-2mm}
\subsubsection{SVERS} To evaluate methods on driving scenarios, we collected a synthetic vehicle-end road scenes (SVERS) dataset from the video game \textit{Grand Theft Auto V}~\cite{gtav} with vehicle-end camera viewpoints. Compared with MVS-SYNTH, SVERS contains a greater variety of driving scenarios, including overpasses, suburban roads, and mountain roads. The dataset includes over 10,000 frames captured with a resolution of 736 x 448. In our experiments, we randomly select 6,372 frames for training and 710 frames for testing. Some samples of SVERS dataset and MVS-SYNTH dataset are shown in Fig.~\ref{fig:dataset}.


\subsection{Training details}
Our model is implemented by Pytorch with Adam optimizer ($\beta_{1}=0.9$, $\beta_{2}=0.999$, $weight\_decay = 10^{-5}$). For indoor dataset (NYUD-V2), we set $init\_lr= 10^{-5}$ while for outdoor datasets (MVS-SYNTH and SVERS), we set $init\_lr= 10^{-4}$.
The learning rate is polynomially decayed with polynomial power of 0.9. The model is trained using a joint loss function that includes depth loss, ASN loss, and normal loss terms, over a period of 20 epochs. The whole training is completed with 6 batches on one A100 GPU. We adopt $5 \times 5$ local patch, 40 sampling triplets, and a sampling ratio of $40\%$ in all experiments. 

\subsection{Evaluation metrics}
To evaluate our method, we compare our method with other SOTA methods in three aspects: the accuracy of depth estimation, the accuracy of surface normal including recovered normal and predicted normal, and the quality of recovered point cloud.

\subsubsection{Depth}  
Following the previous method~\cite{eigen2014depth}, we adopt four metrics to evaluate the predicted depth quantitatively. They are mean absolute relative error (rel), mean log10 error (log10), root mean squared error (rmse), and the accuracy under threshold ($\delta < {1.25}^{i} \text{ where } i\in\{1,2,3\}$).

\vspace{-2mm}
\subsubsection{Surface normal} 
Like prior works~\cite{eigen2015predicting,qi2018geonet}, we evaluate surface normal with the following metrics: the mean of angle error (mean), the median of the angle error (median), and the accuracy below threshold $t \text{ where } t \in\left[11.25^{\circ}, 22.5^{\circ}, 30^{\circ}\right]$.

\vspace{-2mm}
\subsubsection{Point cloud} 
For quantitatively evaluating point clouds converted from predicted depth maps, we use the following metrics: mean Euclidean distance (dist), root mean squared Euclidean distance (rms), and the accuracy below threshold $t \text{ where } t \in\left[0.1\mathrm{m}, 0.3\mathrm{m}, 0.5\mathrm{m}\right]$. We measure distances between a predicted 3D point and its corresponding ground truth point.

\section{Experiments}
\label{exp}
\subsection{Evaluations}

\begin{table}[t]
\setlength{\abovecaptionskip}{0pt}
\setlength{\belowcaptionskip}{0pt}
\begin{center}
\caption{Depth evaluation on NYUD-V2 dataset.
}
\vspace{-2mm}
\label{tab:nyu_depth_eval}
\resizebox{\linewidth}{!}{%
\begin{tabular}{l | c c c c c c c}
\hline
Method & \textbf{rel} ($\downarrow$) & \textbf{log10} ($\downarrow$) & \textbf{rmse} ($\downarrow$) & $ \boldsymbol{\delta_{1}}$ ($\uparrow$) & $ \boldsymbol{\delta_{2}}$ ($\uparrow$) &  $\boldsymbol{\delta_{3}}$ ($\uparrow$)\\
\hline
Saxena~\etal~\cite{saxena2008make3d} & 0.349 & - & 1.214 & 0.447 & 0.745 & 0.897\\ 
Karsch~\etal~\cite{karsch2012depth} & 0.349 & 0.131 & 1.21 & - & - & - \\
Liu~\etal~\cite{liu2014discrete} & 0.335 & 0.127 & 1.06 & - & - & - \\
Ladicky~\etal~\cite{ladicky2014pulling} & - & - & - & 0.542 & 0.829 & 0.941 \\
Li~\etal~\cite{li2015depth} & 0.232 & 0.094 & 0.821 & 0.621 & 0.886 & 0.968 \\
Roy~\etal~\cite{roy2016monocular} & 0.187 & 0.078 & 0.744 & - & - & - \\
Liu~\etal~\cite{liu2015learning} & 0.213 & 0.087 & 0.759 & 0.650 & 0.906 & 0.974 \\
Wang~\etal~\cite{wang2015towards} & 0.220 & 0.094 & 0.745 & 0.605 & 0.890 & 0.970\\
Eigen~\etal~\cite{eigen2015predicting} & 0.158 & - & 0.641 & 0.769 & 0.950 & 0.988 \\
Chakrabarti~\etal~\cite{chakrabarti2016depth} & 0.149 & - & 0.620 & 0.806 & 0.958 & 0.987 \\
Li~\etal~\cite{li2017two} & 0.143 & 0.063 & 0.635 & 0.788 & 0.958 & 0.991 \\
Laina~\etal~\cite{laina2016deeper} & 0.127 & 0.055 & 0.573 & 0.811 & 0.953 & 0.988 \\
Hu~\etal~\cite{hu2019revisiting} & 0.115 & 0.050 & 0.530 & 0.866 & 0.975 & 0.993 \\
DORN~\cite{fu2018deep} & 0.115 & 0.051 & 0.509 & 0.828 & 0.965 & 0.992 \\
GeoNet~\cite{qi2018geonet} & 0.128 &  0.057 & 0.569 & 0.834 & 0.960 & 0.990 \\
GeoNet++~\cite{qi2020geonet++} & 0.113 & 0.049 & 0.527 & 0.862 & 0.965 & 0.989 \\
VNL~\cite{yin2019enforcing} & 0.108 & 0.048 & 0.416 & 0.875 & 0.976 & 0.994 \\
BTS~\cite{lee2019big} & 0.113 & 0.049 & 0.407 & 0.871 & 0.977 & 0.995 \\
ASN~\cite{long2021adaptive} & \textbf{0.101} & 0.044 & 0.377 & 0.890 & 0.982 & 0.996\\
Ours & \textbf{0.101} & \textbf{0.043} & \textbf{0.368} & \textbf{0.906} & \textbf{0.985} & \textbf{0.997}\\
\hline

\hline
\end{tabular}%
}
\end{center}
\vspace{-0.2in}
\end{table}

\begin{table}[!t]
\begin{center}
\caption{Depth evaluation on outdoor datasets.}
\vspace{-5mm}
\label{tab:synthetic_depth_eval}
\resizebox{\linewidth}{!}{%
\begin{tabular}{ l | l | c c c c c c } 
\hline
Dataset & Method & \textbf{rel} ($\downarrow$) & \textbf{log10} ($\downarrow$) & \textbf{rmse} ($\downarrow$) & $ \boldsymbol{\delta_{1}}$ ($\uparrow$) & $ \boldsymbol{\delta_{2}}$ ($\uparrow$) &  $\boldsymbol{\delta_{3}}$ ($\uparrow$)\\
\hline
\multirow{4}{*}{SVERS}  
& BTS~\cite{lee2019big} & 0.093 & 0.034 & 5.240 & 0.939 & 0.977 & 0.987\\
& VNL~\cite{yin2019enforcing} & 0.067 & 0.030 & 4.710 & 0.930 & 0.982 & 0.994 \\
& ASN~\cite{long2021adaptive} & 0.060 & 0.026 & 3.541 & 0.956 & 0.991 & 0.997 \\
& Ours~ & \textbf{0.056} & \textbf{0.024} & \textbf{3.160} & \textbf{0.964} & \textbf{0.993} & \textbf{0.998} \\
\hline
\multirow{4}{*}{MVS-SYNTH}  
& BTS~\cite{lee2019big} & 0.106 & 0.038 & 4.871 & 0.925 & 0.974 & 0.987\\
& VNL~\cite{yin2019enforcing} & 0.101 & 0.037 & 4.253 & 0.916 & 0.977 & 0.991 \\
& ASN~\cite{long2021adaptive} & 0.081 & 0.034 & 3.607 & 0.935 & 0.986 & \textbf{0.995} \\
& Ours~ & \textbf{0.080} & \textbf{0.031} & \textbf{3.259} & \textbf{0.948} & \textbf{0.987} & \textbf{0.995} \\
\hline

\hline
\end{tabular}%
}
\end{center}
\vspace{-0.2in}
\end{table}

\subsubsection{Depth estimation} We compare the proposed method with several SOTA methods that also use geometric constraints, including i) GeoNet++~\cite{qi2020geonet++} (least squares normal); ii) VNL~\cite{yin2019enforcing} (virtual normal constraint); iii) BTS~\cite{lee2019big} (predict local plane equations not directly predict depth); ASN~\cite{long2021adaptive} (local adaptive surface normal constraint) on both indoor and outdoor datasets.

The quantitative comparisons on NYUD-V2 dataset are shown in Table~\ref{tab:nyu_depth_eval}, where our method significantly outperforms other SOTA methods across all evaluation metrics. Additionally, our method's exceptional performance in outdoor scenes is further demonstrated with results on the MVS-SYNTH and SVERS datasets, shown in Table~\ref{tab:synthetic_depth_eval}. Moreover, to further evaluate the generalization of our method, we perform cross-evaluation on unseen ScanNet dataset using the models trained on NYU dataset. As shown in Table~\ref{tab:scannet_depth_eval}, our method still outperforms other methods, especially on the \textbf{rmse} and $ \boldsymbol{\delta_{1}}$ metrics.

We also show the qualitative comparisons on the aforementioned datasets where our method shows the best performance among all methods in terms of the ability to accurately recover the geometries of the scene. 
Fig.~\ref{fig:nyu_compare} showcases the visual results in indoor scenes. It can be seen that our predicted depth map successfully recovers sharp and clear shape details, as evidenced by the clear depiction of the lamps in the second group. 
For outdoor scenes, as shown in Fig.~\ref{fig:SYNTH}, our method still exhibits outstanding performance by leveraging geometric context to produce accurate structures and detailed shape representations, as evidenced by the correct depiction of traffic lights, pedestrians, and buildings.
In addition, we demonstrate the robust generalization capacity of our method with cross-evaluation qualitative results, as shown in Fig.~\ref{fig:scannet}, where the estimated depth maps successfully preserve the geometric details.


\begin{table}[!t]
\begin{center}
\caption{Depth evaluation on ScanNet dataset.
}
\vspace{-5mm}
\label{tab:scannet_depth_eval}
\resizebox{\linewidth}{!}{%
\begin{tabular}{l | c c c c c c } 
\hline
Method & \textbf{rel} ($\downarrow$) & \textbf{log10} ($\downarrow$) & \textbf{rmse} ($\downarrow$) & $ \boldsymbol{\delta_{1}}$ ($\uparrow$) & $ \boldsymbol{\delta_{2}}$ ($\uparrow$) &  $\boldsymbol{\delta_{3}}$ ($\uparrow$)\\
\hline
GeoNet~\cite{qi2018geonet} & 0.255 & 0.106 & 0.519 & 0.561 & 0.855 & 0.958\\
GeoNet++~\cite{qi2020geonet++} & 0.244 & 0.104 & 0.510 & 0.566 & 0.857 & 0.960 \\
VNL~\cite{yin2019enforcing}  & 0.238 & 0.105 & 0.505 & 0.565 & 0.856 & 0.957\\
BTS~\cite{lee2019big} & 0.246 & 0.104 & 0.506 & 0.583 & 0.858 & 0.951\\
ASN~\cite{long2021adaptive} & 0.233 & 0.100 & 0.484 & 0.609 & 0.861 & 0.955 \\
Ours~ & \textbf{0.198} & \textbf{0.077} & \textbf{0.367} & \textbf{0.721} & \textbf{0.929} & \textbf{0.980} \\
\hline

\hline
\end{tabular}%
}
\end{center}
\vspace{-0.2in}
\end{table}

\begin{table}[t]
\begin{center}
\caption{Point cloud evaluation on NYUD-V2 dataset.
}
\vspace{-5mm}
\label{tab:point_cloud_eval_nyud}
\resizebox{\linewidth}{!}{%
\begin{tabular}{l | c c c c c } 
\hline
Method & \textbf{dist} ($\downarrow$) & \textbf{rmse} ($\downarrow$) & $\boldsymbol{0.1m}$ ($\uparrow$) & $\boldsymbol{0.3m}$ ($\uparrow$) & $\boldsymbol{0.5m}$ ($\uparrow$) \\
\hline
VNL~\cite{yin2019enforcing} & 0.515 & 0.686 & 0.181 & 0.469 & 0.644 \\
GeoNet~\cite{qi2018geonet} & 0.399 & 0.615 & 0.217 & 0.549 & 0.739 \\
GeoNet++~\cite{qi2020geonet++} & 0.392 & 0.608 & 0.220 & 0.558 & 0.747 \\    
BTS~\cite{lee2019big}& 0.317 & 0.544 & 0.278 & 0.653 & 0.822 \\
Hu~\etal~\cite{hu2019revisiting} & 0.311 & 0.537 & 0.288 & 0.666 & 0.831\\
ASN~\cite{long2021adaptive} & 0.266 & 0.497 & \textbf{0.332} & \textbf{0.727} & 0.869\\
Ours &  \textbf{0.258} & \textbf{0.490} & 0.323 & 0.721 & \textbf{0.874}\\
\hline

\hline
\end{tabular}%
}
\end{center}
\vspace{-0.2in}
\end{table}

\vspace{-1mm}
\subsubsection{Point cloud}
Regarding the quality of the reconstructed point cloud, as shown in Table~\ref{tab:point_cloud_eval_nyud}, our baseline, ASN~\cite{long2021adaptive}, outperforms other methods by a significant margin while our method makes some improvement on this basis. Surprisingly, although VNL~\cite{yin2019enforcing} has better performance than GeoNet~\cite{qi2018geonet} in terms of depth evaluation errors, its mean Eucleadian distance inferior to that of GeoNet, highlighting the necessity of evaluation specially designed for point cloud. 

To visually show the geometric properties of the reconstructed point cloud, we present several qualitative results in Fig.~\ref{fig:nyu_compare} where our point cloud has less distortion and is much more accurate. In contrast, the point cloud generated from the depth maps of other methods suffers from severe distortion and struggles to preserve prominent geometric features, such as planes (e.g., walls) and surfaces with high curvatures (e.g., dolls and pillows). We also show a qualitative comparison between our reconstructed point cloud and the ground truth from different views in Fig.~\ref{fig:nyu_compare}. The highly consistent result further demonstrates the superior performance of our proposed method in terms of the quality of 3D geometry. In addition, we demonstrate the generalization ability of our approach by visualizing our reconstructed point cloud on ScanNet dataset, as shown in \ref{fig:scannet}. Despite some deviations in scale, our method is able to accurately recover the shapes of various objects, including washing sinks, chairs, and sofas.
\vspace{-1mm}
\subsubsection{Surface Normal}
We report results for both recovered surface normals and predicted surface normals. As shown in Table~\ref{tab:nyu_normal_eval} and Table~\ref{tab:synth_normal_eval}, our recovered surface normals have considerably better quality than those produced by other methods on both indoor and outdoor datasets. What’s even more surprising is that the accuracy of our recovered surface normal is even higher than most methods that explicitly predict surface normals. For predicted surface normals, our method outperforms other normal estimation algorithms by a large margin on NYUD-V2 dataset, as shown in \textit{Predicted Surface Normal from the Network} of Table~\ref{tab:nyu_normal_eval}. Our method also performs impressively on outdoor datasets, as indicated in  Table~\ref{tab:synth_normal_eval}.

The qualitative comparisons of indoor and outdoor scenes are presented in Fig.~\ref{fig:nyu_compare} and Fig.~\ref{fig:SYNTH}, respectively. It can be seen that our recovered surface normal is smoother and more accurate than those produced by other methods. This highlights the effectiveness of our strategy in correlating normal constraints with depth estimation, enabling accurate depth estimation along with reliable surface normal and 3D geometry. On the other hand, even on the outdoor datasets, our method not only accurately predicts normals of planes, but also faithfully recovers the details, such as license plates, small windows, and leaves.  Moreover, we visualize the recovered normal and predicted normal on the unseen ScanNet dataset in Fig.~\ref{fig:scannet}. The clear geometric boundaries and high quality of the recovered and predicted surface normals demonstrate the impressive generalization ability of our method.

\begin{table}[!t]
\begin{center}
\caption{Surface normal evaluation on NYUD-V2 dataset.
}
\vspace{-4mm}
\label{tab:nyu_normal_eval}
\resizebox{\linewidth}{!}{%
\begin{tabular}{l | c c c c c}
\hline
Method & \textbf{Mean} ($\downarrow$) & \textbf{Median} ($\downarrow$) & $\boldsymbol{11.25^{\circ}}$ ($\uparrow$) & $\boldsymbol{22.5^{\circ}}$ ($\uparrow$) & $\boldsymbol{30^{\circ}}$ ($\uparrow$)\\
\hline
\hline
\multicolumn{6}{c}{Predicted Surface Normal from the Network} \\
\hline
\hline
3DP~\cite{fouhey2013data} & 33.0 & 28.3 & 18.8 & 40.7 & 52.4 \\
Ladicky~\etal~\cite{ladicky2014pulling} & 35.5 & 25.5 & 24.0 & 45.6 & 55.9 \\
Fouhey~\etal~\cite{fouhey2014unfolding} & 35.2 & 17.9 & 40.5 & 54.1 & 58.9 \\
Wang~\etal~\cite{wang2015designing} & 28.8 & 17.9 & 35.2 & 57.1 & 65.5 \\
Eigen~\etal~\cite{eigen2015predicting} & 23.7 & 15.5 & 39.2 & 62.0 & 71.1 \\
GeoNet~\cite{qi2018geonet} & 19.0 & 11.8 & 48.4 & 71.5 & 79.5 \\
GeoNet++~\cite{qi2020geonet++} & 18.5 & 11.2 & 50.2 & 73.2 & 80.7 \\
Ours & \textbf{15.3} & \textbf{8.0} & \textbf{61.0} & \textbf{78.8} & \textbf{84.9} \\
\hline
\hline
\multicolumn{6}{c}{Recovered Surface Normal from the Point cloud} \\
\hline
\hline
BTS~\cite{lee2019big} & 44.0 & 35.4 & 14.4 & 32.5 & 43.2 \\
GeoNet~\cite{qi2018geonet} & 36.8 & 32.1 & 15.0 & 34.5 & 46.7 \\
DORN~\cite{fu2018deep} & 36.6 & 31.1 & 15.7 & 36.5 & 49.4  \\
Hu~\etal~\cite{hu2019revisiting} & 32.1 & 23.5 & 24.7 & 48.4 & 59.9 \\
GeoNet++~\cite{qi2020geonet++} & 28.3 & 19.9 & 30.1 & 54.7 & 65.5 \\
VNL~\cite{yin2019enforcing} & 24.6 & 17.9 & 34.1 & 60.7 & 71.7  \\
ASN~\cite{long2021adaptive} & 20.0 & 13.4 & 43.5 & 69.1 & 78.6 \\
Ours & \textbf{18.2} & \textbf{11.5} & \textbf{49.2} & \textbf{73.8} & \textbf{81.9} \\
\hline

\hline
\end{tabular}%
}
\end{center}
\vspace{-0.2in}
\end{table}

\begin{table}[!t]
\begin{center}
\caption{Surface normal evaluation on outdoor datasets.
}
\vspace{-4mm}
\label{tab:synth_normal_eval}
\resizebox{\linewidth}{!}{%
\begin{tabular}{l | l | c c c c c}
\hline
Dataset & Method & \textbf{Mean} ($\downarrow$) & \textbf{Median} ($\downarrow$) & $\boldsymbol{11.25^{\circ}}$ ($\uparrow$) & $\boldsymbol{22.5^{\circ}}$ ($\uparrow$) & $\boldsymbol{30^{\circ}}$ ($\uparrow$)\\
\hline
\hline
\multicolumn{7}{c}{Predicted Surface Normal from the Network} \\
\hline
\hline
SVERS & \multirow{2}{*}{ASN++} & 10.4 & 1.39 & 82.7 & 87.3 & 89.0  \\
MVS-SYNTH &  & 14.1 & 3.36 & 73.9 & 82.9 & 86.0\\
\hline
\hline
\multicolumn{7}{c}{Recovered Surface Normal from the Point cloud} \\
\hline
\hline
\multirow{4}{*}{SVERS}  
& BTS~\cite{lee2019big} & 34.7 & 19.6 & 35.1 & 53.6 & 61.2\\
& VNL~\cite{yin2019enforcing} & 24.8 & 13.9 & 48.3 & 69.5 & 78.6 \\
& ASN~\cite{long2021adaptive} & 20.4 & 9.6 & 57.0 & 77.0 & 81.2 \\
& Ours & \textbf{18.5} & \textbf{8.8} & \textbf{61.2} & \textbf{79.7} & \textbf{83.6}  \\
\hline
\multirow{4}{*}{MVS-SYNTH}  
& BTS~\cite{lee2019big} & 43.2 & 30.2 & 25.0 & 41.7 & 49.8 \\
& VNL~\cite{yin2019enforcing} & 30.3 & 19.2 & 31.1 & 60.8 & 69.4 \\
& ASN~\cite{long2021adaptive} & 25.5 & 14.8 & 37.7 & 65.3 & 73.5 \\
& Ours & \textbf{22.5} & \textbf{12.3} & \textbf{46.1} & \textbf{70.4} & \textbf{77.6} \\
\hline

\hline
\end{tabular}%
}
\end{center}
\vspace{-0.2in}
\end{table} 
\section{Discussions}
\label{discuss}

In this section, we further conduct a series of evaluations to give more insights into the proposed method. 

\vspace{-2mm}
\subsection{Ablation study of ASN}
To deeply analyze our proposed Adaptive Surface Normal (ASN) constraint, we leverage \textit{HRNet-18}~\cite{wang2020deep} as backbone but without the normal prediction layer
and conduct experiments in this part.

\subsubsection{Effectiveness} To validate the effectiveness of our proposed adaptive surface normal constraint, we train models with different constraints: a) only $L_1$ depth constraint; b) depth and Sobel-like operator surface normal constraints (SOSN); c) depth and least square surface normal constraints (LSSN); d) depth and virtual normal constraints (VN); e) depth and our adaptive surface normal constraints (ASN).

\begin{figure}[!t]
\setlength{\abovecaptionskip}{1pt}
\setlength{\belowcaptionskip}{1pt}
    \centering
    \includegraphics[width=\linewidth]{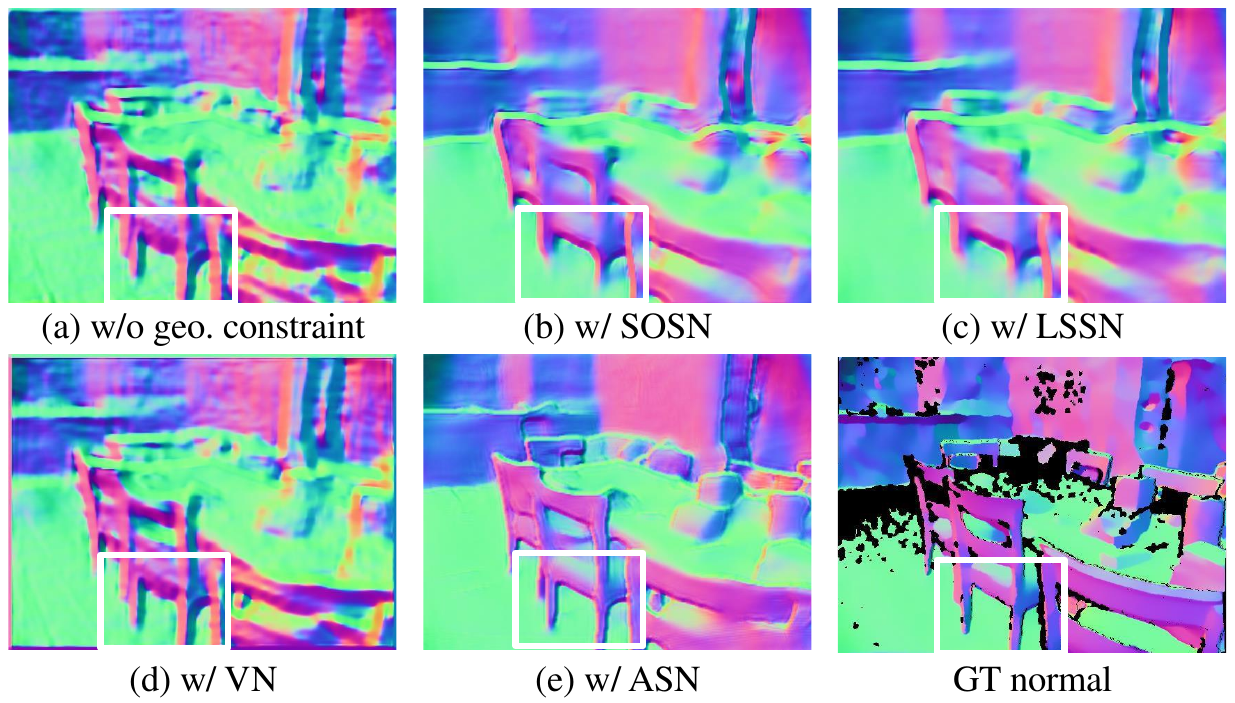}
    \caption{Comparisons of models with different geometric constraints. Model with our ASN constraint achieves better surface normal estimation, even accurately capture detailed geometries, like the legs of chairs (see white boxes).}
    \label{fig:different_constraints}
    \vspace{-3mm}
\end{figure}

As shown in Table~\ref{tab:diff_constraints}, the model with our adaptive surface normal constraint (ASN) outperforms all the others. Although the models with Sobel-like operator (SOSN) and least square normal constraint (LSSN) show better recovered surface normal, their depth estimation accuracy drops off compared with the model without geometric constraints. The model with virtual normal (VN)~\cite{yin2019enforcing} constraint shows the worst quality of recovered surface normal among the four types of geometric constraints, given that virtual normal is derived from global sampling on the estimated depth map, which inevitably loses local geometric information.  
Furthermore, we give a set of qualitative comparisons in Fig.~\ref{fig:different_constraints}. The results clearly show our ASN constraint achieves better surface normal estimation results and captures detailed geometric features, even for the thin structures like the legs of chairs.

\begin{table}[!t]
\begin{center}
\caption{Comparisons of models with different geometric constraints on NYUD-V2 dataset.
}
\vspace{-5mm}
\label{tab:diff_constraints}
\resizebox{\linewidth}{!}{%
\begin{tabular}{l | c c c | c c c }
\hline
\multirow{2}{*}{Constraints} & \textbf{rel} ($\downarrow$) & \textbf{log10} ($\downarrow$) & $\boldsymbol{\delta_{1}}$ ($\uparrow$) & \textbf{Mean} ($\downarrow$) & \textbf{Median} ($\downarrow$) & $\boldsymbol{11.25^{\circ}}$ ($\uparrow$) \\
& \multicolumn{3}{c}{Depth} & \multicolumn{3}{|c}{Recovered normal} \\
\hline
L1 & 0.113 & 0.047  & 0.875   & 31.3 & 23.2 & 24.9 \\
L1 + SOSN & 0.118 & 0.049 & 0.867  & 22.8 & 16.1 & 36.2 \\
L1 + LSSN & 0.119 & 0.050  & 0.862  & 23.5 & 16.3 & 35.7 \\
L1 + VN & 0.111 & 0.047  & 0.876  & 31.7 & 21.4 & 28.4 \\
L1 + ASN & \textbf{0.111} & \textbf{0.047} & \textbf{0.876} & \textbf{22.2} & \textbf{15.8} & \textbf{36.9} \\
\hline

\hline
\end{tabular}%
}
\end{center}
\vspace{-3mm}
\end{table}

\subsubsection{Adaptive modules}
To evaluate the effect of the proposed two adaptive modules, i.e., geometric context adaption and area adaption, we conduct an ablation study. We train models with different adaptive modules: only Geometric Context (GC) adaption, only Area adaption, and both. From Table~\ref{tab:abalation}, we can see the model with both adaptive modules achieves the best performance, which verifies the necessity of each adaptive module. 

\begin{table}[!t]
\begin{center}
\caption{Ablation study of the proposed adaptive modules on NYUD-V2 dataset. We evaluate the accuracy of the recovered surface normals.}
\vspace{-5mm}
\label{tab:abalation}
\resizebox{\linewidth}{!}{%
\begin{tabular}{l | c c c c c }
\hline
Module & \textbf{Mean} ($\downarrow$) & \textbf{Median} ($\downarrow$) & $\boldsymbol{11.25^{\circ}}$ ($\uparrow$) & $\boldsymbol{22.5^{\circ}}$ ($\uparrow$) & $\boldsymbol{30^{\circ}}$ ($\uparrow$)\\
\hline
only Area &22.6 & 16.0 & 36.4 & 63.6 & 74.4 \\
only GC & 22.3 & \textbf{15.8} & \textbf{36.9} & 64.1 & 74.8\\
Area+GC & \textbf{22.2} & \textbf{15.8} & \textbf{36.9} & \textbf{64.2} & \textbf{74.9} \\
\hline

\hline
\end{tabular}%
}
\end{center}
\vspace{-0.2in}
\end{table}

\begin{figure}[!t]
\setlength{\abovecaptionskip}{1pt}
\setlength{\belowcaptionskip}{1pt}
    \centering
    \includegraphics[width=\linewidth]{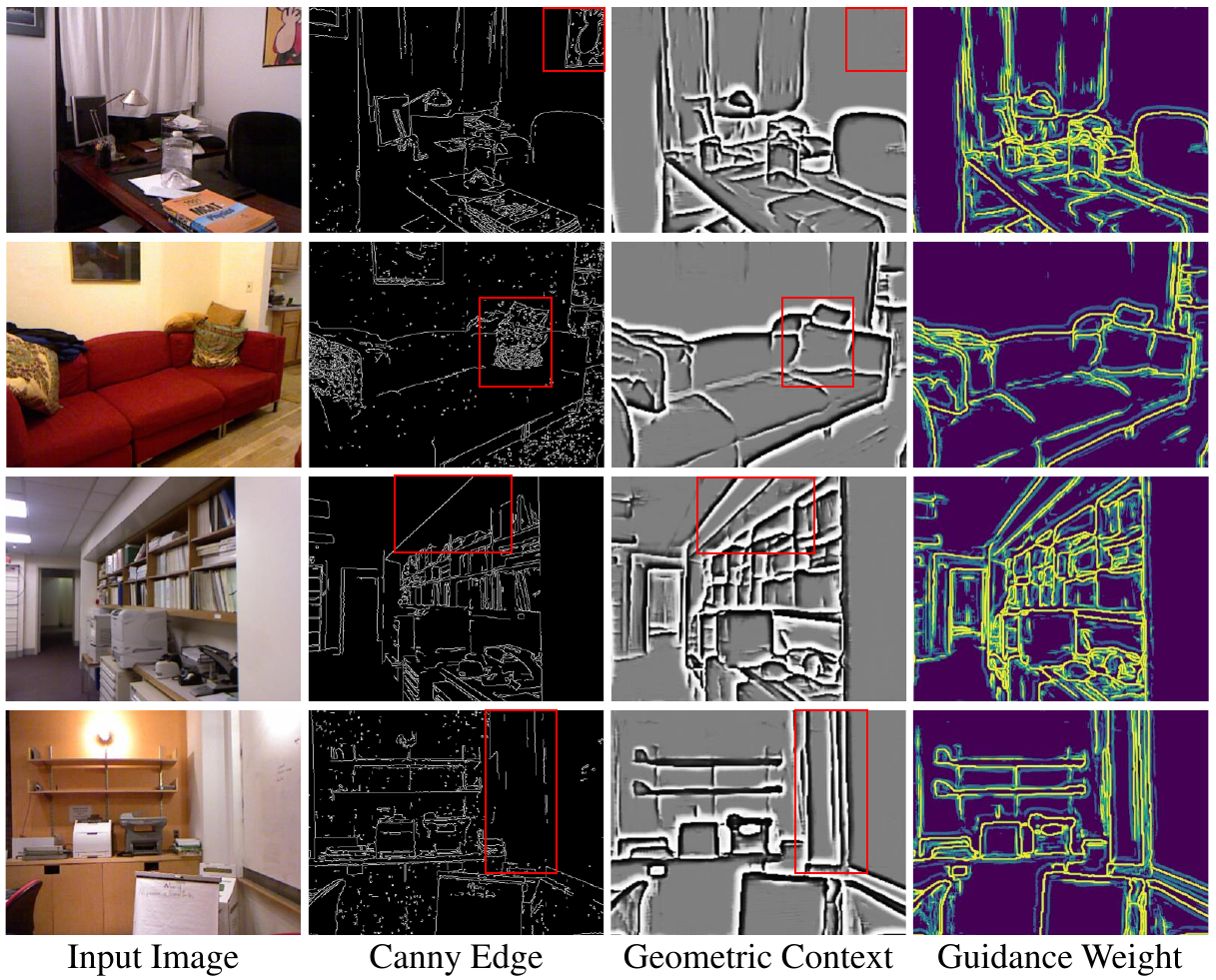}
    \caption{Visualization of our geometric context and guidance weight maps (first-order derivatives). We visualize three-channel geometric context, which can clearly show the geometric context. While the guidance weight maps are presented in one channel to depict the exact locations of geometric edges. For comparison, we also visualize the edge maps detected by the Canny operator. }
    \label{fig:guidance_feature_vis}
    \vspace{-3mm}
\end{figure}

\subsubsection{Visualization of geometric context} 
\label{context_feature}
The geometric adaptive module is the key to our adaptive surface normal constraint method. To better understand what the network learns, we visualize the learned features of the geometric context. We plot one from the multiple channels of the geometric context, which is shown in Fig.~\ref{fig:guidance_feature_vis}. It can be seen that the learned geometric context comprehensively captures the geometric details such as geometric shapes and small structures, thereby providing informative boundaries.

For comparison, we use the Canny operator to detect the edges of the input image by image intensity variance. As we can see, our geometric context is not simply coincident with the Canny edges. For example, the Canny operator detects fragmented edges based on the texture of the mural and cushion, while our geometric context indicates the true shape boundaries where the 3D geometry changes. Besides, Canny operator may fail in detecting geometric details such as intersection lines of walls. In contrast, our geometric context can faithfully capture these details.

We also visualize the guidance weight, as shown in the fourth column of Fig.~\ref{fig:guidance_feature_vis}. Compared to the geometric context, the guidance weight can indicate the exact locations of the geometric details. 
As you can see, the brighter pixels of the guidance weight map indicate the locations where the geometry changes.
As a result, we utilize the guidance weight to sample pixels for normal estimation.

\begin{figure}[t]
\setlength{\abovecaptionskip}{1pt}
\setlength{\belowcaptionskip}{1pt}
    \centering
    \includegraphics[width=\linewidth]{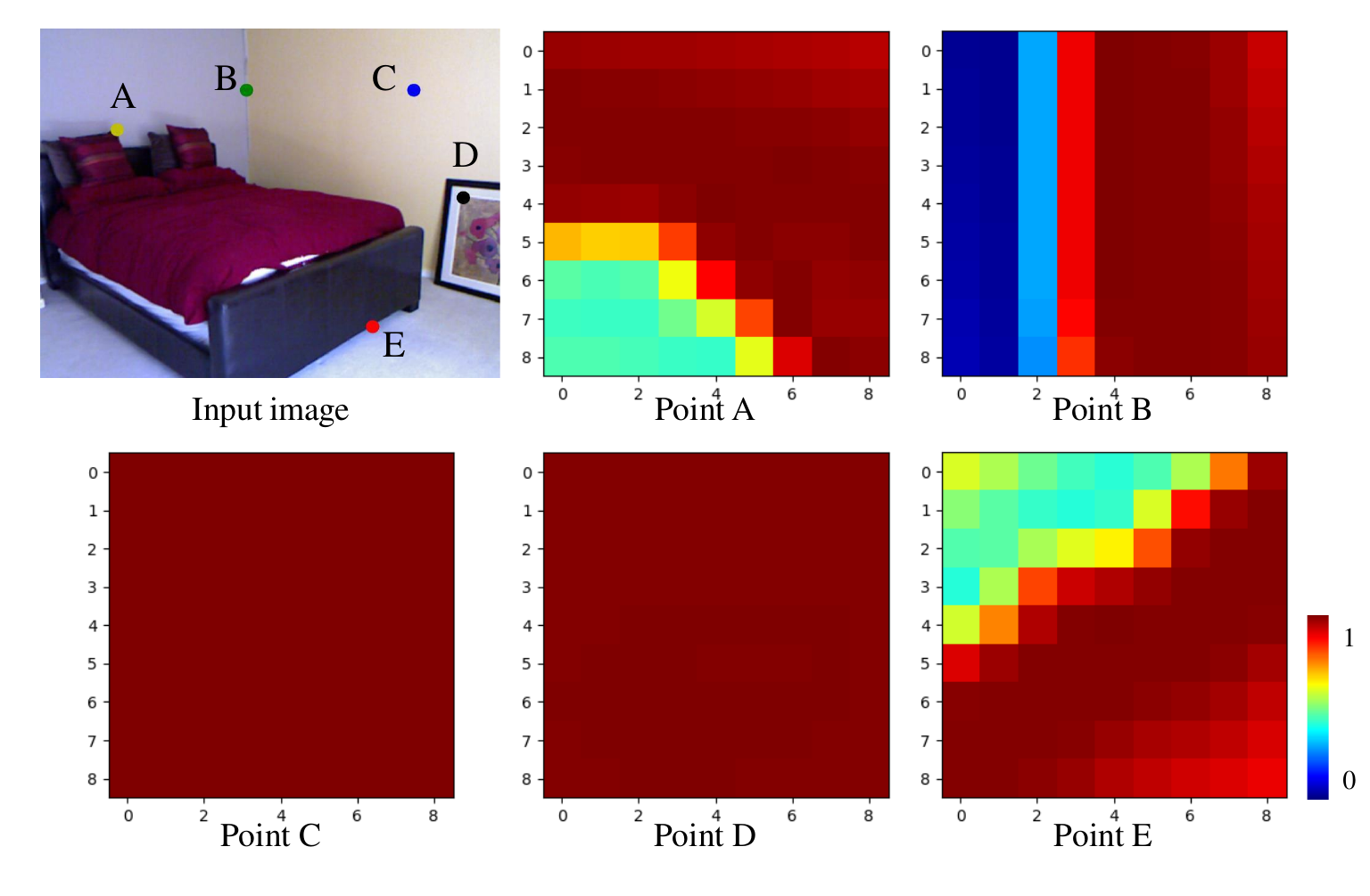}
    \caption{The visualization of similarity kernels. 
    The similarity kernels of Point A, B, and E demonstrate that our method could successfully distinguish different geometries. The similarity kernels of Point C and D further show that our method captures the 3D geometry variances of the shapes in the 3D world, instead of the image color distinctions. 
    }
    \label{fig:kernel_vis}
    \vspace{-2mm}
\end{figure}

\begin{figure}[!t]
\setlength{\abovecaptionskip}{1pt}
\setlength{\belowcaptionskip}{1pt}
    \centering
    \includegraphics[width=\linewidth]{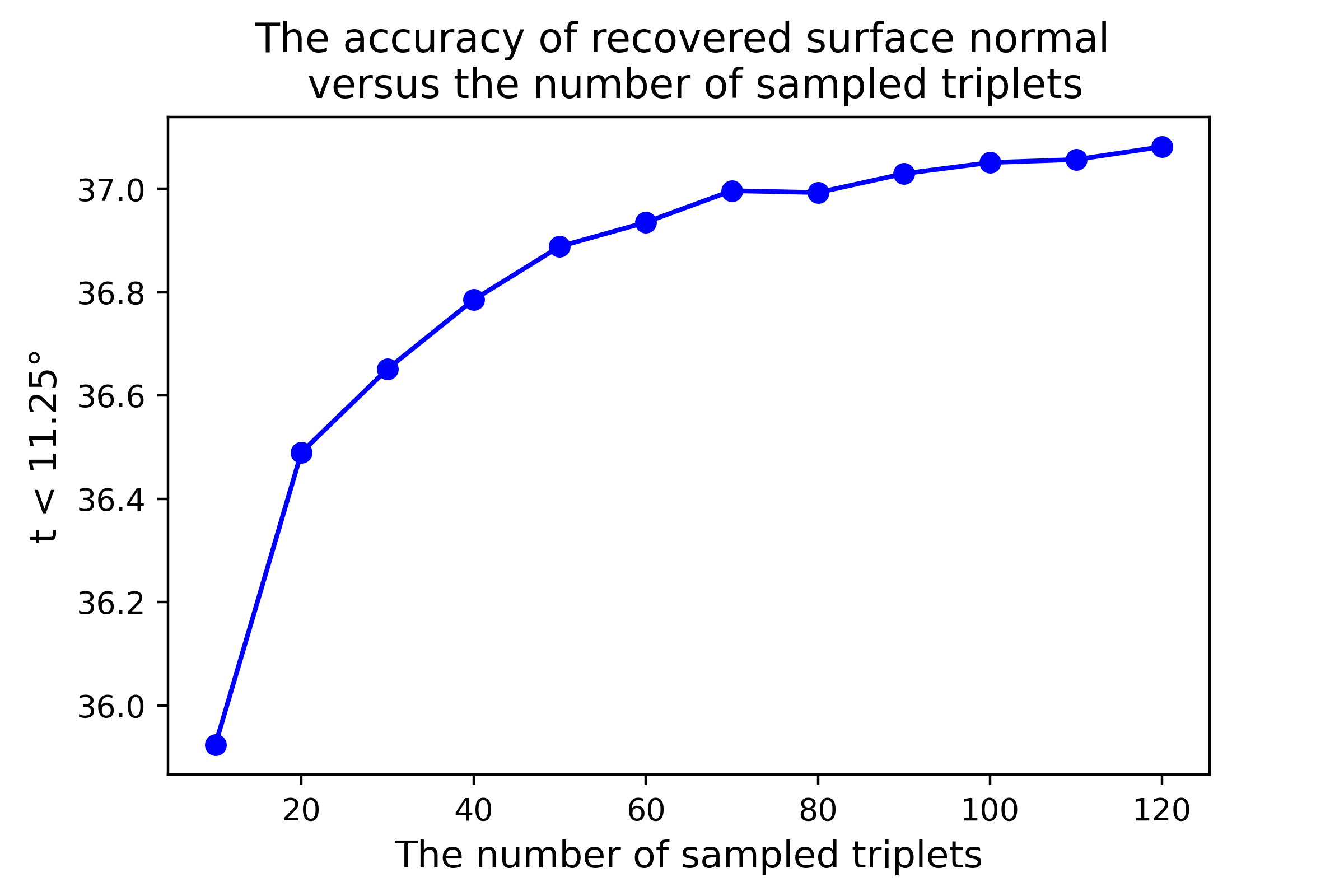}
    \caption{The accuracy of recovered surface normal versus the number of sampled triplets. The more triplets are sampled, the more accurate the recovered surface normal is.}
    \label{fig:num_of_triplets}
\end{figure}

\subsubsection{Visualization of similarity kernels}
To validate whether our model can capture the true geometric boundaries of shapes, we select five points on the image and visualize their color-coded similarity kernels in Fig.~\ref{fig:kernel_vis}. The similarity kernels of Point A, B, and E indicate that our method could successfully distinguish different geometries, such as shape boundaries and corners. Furthermore, the similarity kernels of Point C and D show that our approach captures the 3D geometry variances of the shapes in the real world, instead of the color distinctions of the image. For example,  Point D has large color variances in the image, but its similarity kernel has constant values indicating the unchanged geometry.

\subsubsection{Number of sampled triplets}
To quantitatively analyze the influence of the number of sampled triplets, we recover surface normals from our estimated depth maps using our adaptive surface normal computation method with $5 \times 5$ local patch.
Based on Fig.~\ref{fig:num_of_triplets}, it is not surprised that more sampled triplets will contribute to more accurate surface normals. The accuracy increases dramatically from $10 \sim 20$ sampled triplets and gradually saturates with more triplets sampled. To balance efficiency and accuracy, the number of sampled triplets is recommended to be $40 \sim 60$.

\begin{figure}[!t]
\setlength{\abovecaptionskip}{1pt}
\setlength{\belowcaptionskip}{1pt}
    \centering
    \includegraphics[width=\linewidth]{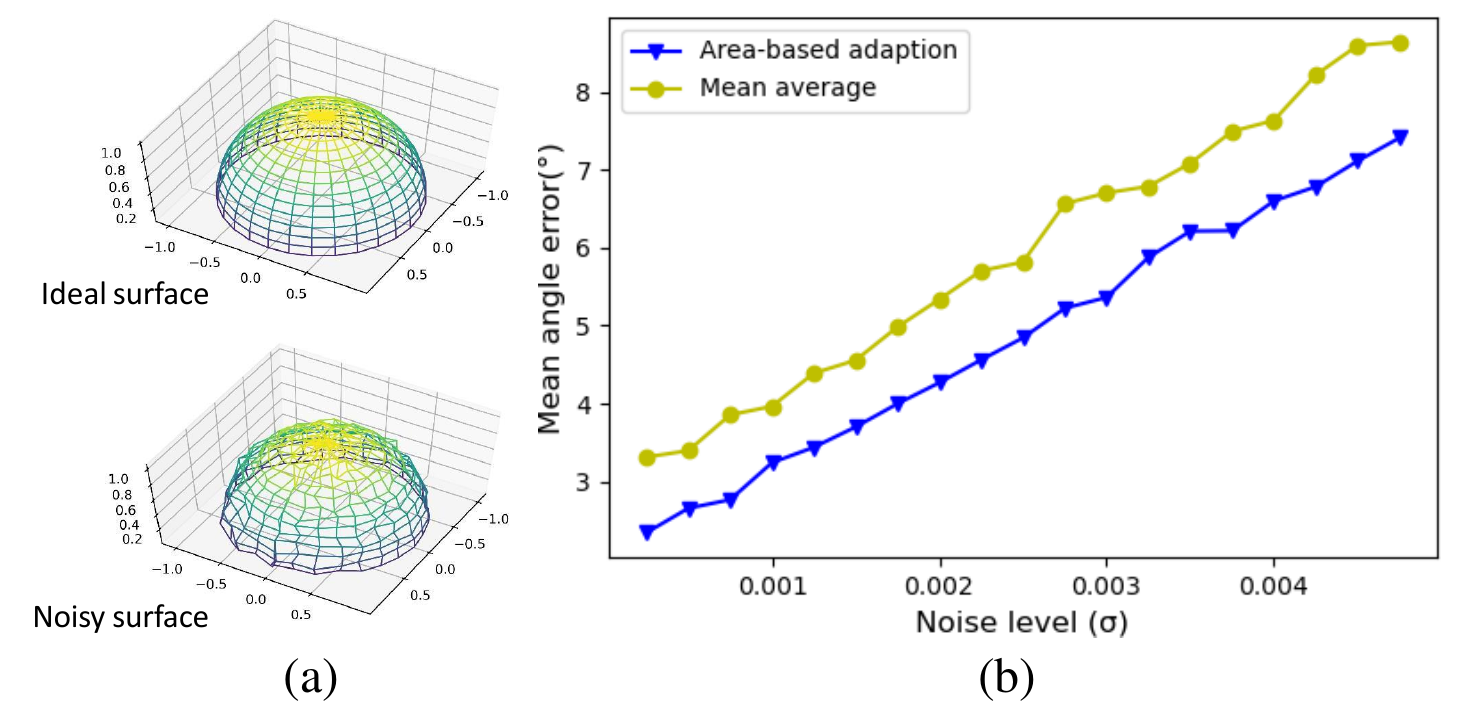}
    \caption{Effectiveness of the area-based adaption. (a) The ideal and noisy surface. (b) We employ the mean angle error to evaluate surface normals estimated by the simple average strategy and our area-based adaption. Compared with the simple average strategy, our area-based adaption is more robust to noises.}
    \label{fig:noise_exp}
    \vspace{-2mm}
\end{figure}

\begin{figure}[!t]
\setlength{\abovecaptionskip}{1pt}
\setlength{\belowcaptionskip}{1pt}
    \centering
    \includegraphics[width=\linewidth]{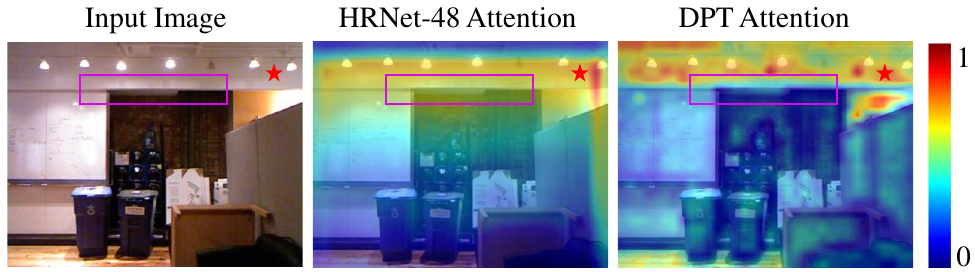}
    \caption{Self-attention maps for the queried pixel (marked as a red star). The attention map of DPT shows sharper boundaries that validate the superior ability of DPT to capture structural information.
    }
    \label{fig:attn_map}
    \vspace{-1mm}
\end{figure}

\subsubsection{Size of local patch}
We evaluate the effect of the size of local patch (window size) to our method by training the network with different local patch sizes. As illustrated in Table~\ref{tab:patch_size}, a larger local patch could improve the performance, especially for the surface normal, but the improvements are not significant. The reason behind is, our ASN constraint is an adaptive strategy that can automatically determine the reliability of the sampled points given different local patches; therefore, our method is robust to the choice of local patch size. 

Additionally, given the image resolution, selecting an appropriate window size is crucial for training the network as it allows for a balance between performance and time efficiency. To find the optimal window size of varying image resolution in both indoor scenes and outdoor scenes, we conducted experiments to analyze the choices of window size and found such rule: the ratio of window area to image area is around $5e^{-5} \sim 1e^{-4} $ in both indoor scenes and outdoor scenes. For example, given an image resolution of $1280\times960$, assuming the ratio is $8e^{-5}$ would suggest a window size of $ \sqrt{1280\times960 \times 8e^{-5}}=9.9$. Since the window size should be an odd integer, it can be set to $9 \times 9$.  More detailed experiments can be seen in Section 1 of the Appendix.

\subsubsection{Area-based adaption} We use the area of a sampled triangle as the combinational weight for adaption. To evaluate the effectiveness of the area-based adaption, we conduct an experiment with a comparison to the simple average strategy. We create a unit semi-sphere surface as noise-free data and then add Gaussian noises to simulate real noisy data (see Fig.~\ref{fig:noise_exp} (a)). We compare the mean of angle errors of the normals estimated by these two methods with the increase of noises, and the results are given in Fig.~\ref{fig:noise_exp} (b). We can see that our area-based adaption gives lower estimation error with the increase of noise level, demonstrating the robustness of the use of the area for adaption.

\begin{figure*}[t]
\setlength{\abovecaptionskip}{1pt}
\setlength{\belowcaptionskip}{1pt}
    \centering
    \includegraphics[width=\linewidth]{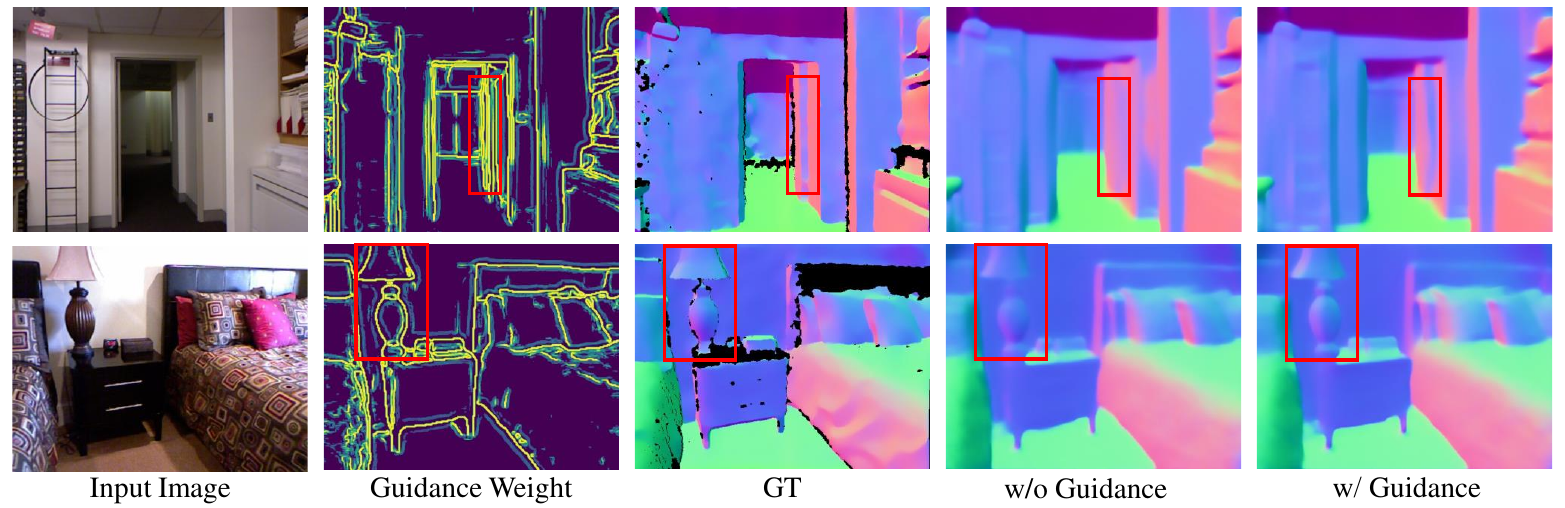}
    \caption{Qualitative comparisons of the predicted surface normal with guidance obtained from the first-order derivatives of geometric context and without the guidance of geometric context. As shown in red bounding boxes, the geometric context effectively improves the quality of predicted normal, especially in regions with rich geometric details.}
    \label{fig:normal_ablation}
    \vspace{-3mm}
\end{figure*}

\begin{table}[!t]
\begin{center}
\caption{The influence of local patch size.
}
\vspace{-2mm}
\label{tab:patch_size}
\resizebox{0.95\linewidth}{!}{%
\begin{tabular}{c | c c c  | c c c}
\hline
\multirow{2}{*}{Size} & \textbf{rel} ($\downarrow$) & \textbf{log10} ($\downarrow$) & $\boldsymbol{\delta_{1}}$ ($\uparrow$) & \textbf{Mean} ($\downarrow$) & \textbf{Median} ($\downarrow$) & $\boldsymbol{11.25^{\circ}}$ ($\uparrow$) \\
& \multicolumn{3}{c}{Depth} & \multicolumn{3}{|c}{ Recovered Normal} \\
\hline
3 & 0.112 & 0.047 & \textbf{0.877} & 22.5 & 15.8 & 36.9\\
5 & \textbf{0.111} & 0.047 & 0.876 & 22.4 & 15.8 & \textbf{37.1}\\
7 & 0.112 & 0.047 & \textbf{0.877} & \textbf{22.2} & \textbf{15.7} & \textbf{37.1}\\
9 & \textbf{0.111} & 0.047 & 0.875 & 22.4 & 15.8 & 37.0 \\
\hline
\end{tabular}%
}
\end{center}
\vspace{-5mm}
\end{table}

\begin{table}[!t]
\begin{center}
\caption{ASN constraint with different backbones for depth estimation.}
\vspace{-3mm}
\label{tab:transformer}
\resizebox{\linewidth}{!}{%
\begin{tabular}{l | c c c | c c c }
\hline
\multirow{2}{*}{Backbone} & \textbf{rel} ($\downarrow$) & \textbf{rmse} ($\downarrow$) & $\boldsymbol{\delta_{1}}$ ($\uparrow$) & \textbf{Mean} ($\downarrow$) & \textbf{Median} ($\downarrow$) & $\boldsymbol{11.25^{\circ}}$ ($\uparrow$) \\
& \multicolumn{3}{c}{Depth} & \multicolumn{3}{|c}{Recovered normal} \\
\hline
HRNet-18 & 0.111 & 0.394 & 0.876 & 22.2 & 15.8 & 37.1 \\
HRNet-48 & 0.101 & 0.377  & 0.890 & 20.0 & 13.4 & \textbf{43.5} \\
DPT & \textbf{0.100} & \textbf{0.369} & \textbf{0.908} & \textbf{19.9} & \textbf{13.3} & 43.4 \\
\hline

\hline
\end{tabular}%
}
\end{center}
\vspace{-5mm}
\end{table}

\vspace{-1mm}
\subsubsection{Time complexity}
Here, we discuss the time complexity of different normal computation methods, including our sampling based method, Sobel-like operator \cite{hu2019revisiting, kusupati2020normal} and least square based method \cite{qi2018geonet,long2020occlusion}.
Ours and the Sobel-like operator only involve matrix addition and vector dot/cross production operations; thus it is easy to show the time complexity is $O\left(n \right)$, while our time complexity will increase linearly with more samples. 
However, the least square module \cite{qi2018geonet,long2020occlusion} directly calculates the closed form solution of least square equations, which involves matrix multiplication, inversion and determinant, leading to the time complexity of $O\left(n^{3} \right)$. Therefore, our method effectively balances the accuracy and computational efficiency.

\subsubsection{Backbone options}
Here we conduct experiments to analyze the effects of different backbones. The quantitative results shown in Table~\ref{tab:transformer} validate that DPT achieves the best performance in terms of estimated depth and the recovered surface normal. 
To better understand the advantages of the DPT backbone, we visualize the self-attention maps of HRNet-48 and DPT for better analysis. As shown in the marked pink bounding boxes in Fig.~\ref{fig:attn_map}, pixels located at or out of the wall exhibit significantly different attention patterns in DPT's attention map compared to the queried pixel (marked as red star). On the other side, this contrast is not noticeable in HRNet-48's attention map. 
The observations can prove that transformer has more powerful ability to capture distinguishable features compared with CNN backbones.



\begin{table}[!t]
\begin{center}
\caption{Ablation study of the geometric context guided normal estimation on NYUD-V2 and SEVERS datasets. We evaluate the accuracy of the predicted surface normals.}
\vspace{-4mm}
\label{tab:refinement}
\resizebox{\linewidth}{!}{%
\begin{tabular}{l | c | c c c }
\hline
\textbf{Dataset} & \textbf{Guidance Calculation} & \textbf{Mean} ($\downarrow$) & \textbf{Median} ($\downarrow$) & $\boldsymbol{11.25^{\circ}}$ ($\uparrow$) \\
\hline
\multirow{3}{*}{NYUD-V2} 
 & w/o guidance & 15.40 & 8.24 & 60.4\\
 & 1st-order & \textbf{15.29} & \textbf{7.97} & \textbf{61.0}\\
 & 2nd-order & 15.30 & 8.00 & 60.9\\
\hline
\hline
\multirow{3}{*}{SEVERS} 
 & w/o guidance & 10.80 & 1.84 & 82.3\\
 & 1st-order  & 10.41 & 1.39 & 82.7\\
 & 2nd-order & \textbf{10.34} & \textbf{1.32} & \textbf{82.9}\\
\hline

\hline
\end{tabular}%
}
\end{center}
\vspace{-5mm}
\end{table}

\begin{table}[!t]
\begin{center}
\caption{Experiments with or without NPL on NYUD-V2 dataset. We evaluate the estimated depth and its recovered surface normal.}
\vspace{-5mm}
\label{tab:guidance_normal_decoder}
\resizebox{\linewidth}{!}{%
\begin{tabular}{l | c c | c c | c c }
\hline
\multirow{2}{*}{Method} & \textbf{rmse} ($\downarrow$) & $\boldsymbol{\delta_{1}}$ ($\uparrow$) & \textbf{Mean} ($\downarrow$) & $\boldsymbol{11.25^{\circ}}$ ($\uparrow$)   & \textbf{rmse} ($\downarrow$) & $\boldsymbol{0.1m}$ ($\uparrow$) \\
& \multicolumn{2}{c}{Depth} & \multicolumn{2}{|c}{Recovered normal} & \multicolumn{2}{|c}{Point Cloud} \\
\hline
w/o NPL & 0.369  & \textbf{0.908}   & 19.9 & 43.4 & 0.508 & 0.317 \\
w/ NPL & \textbf{0.368} & 0.906 & \textbf{18.2} & \textbf{49.2}  & \textbf{0.490} & \textbf{0.322}\\
\hline

\hline
\end{tabular}%
}
\end{center}
\vspace{-3mm}
\end{table}

\begin{figure*}[t]
\setlength{\abovecaptionskip}{1pt}
\setlength{\belowcaptionskip}{1pt}
    \centering
    \includegraphics[width=\linewidth]{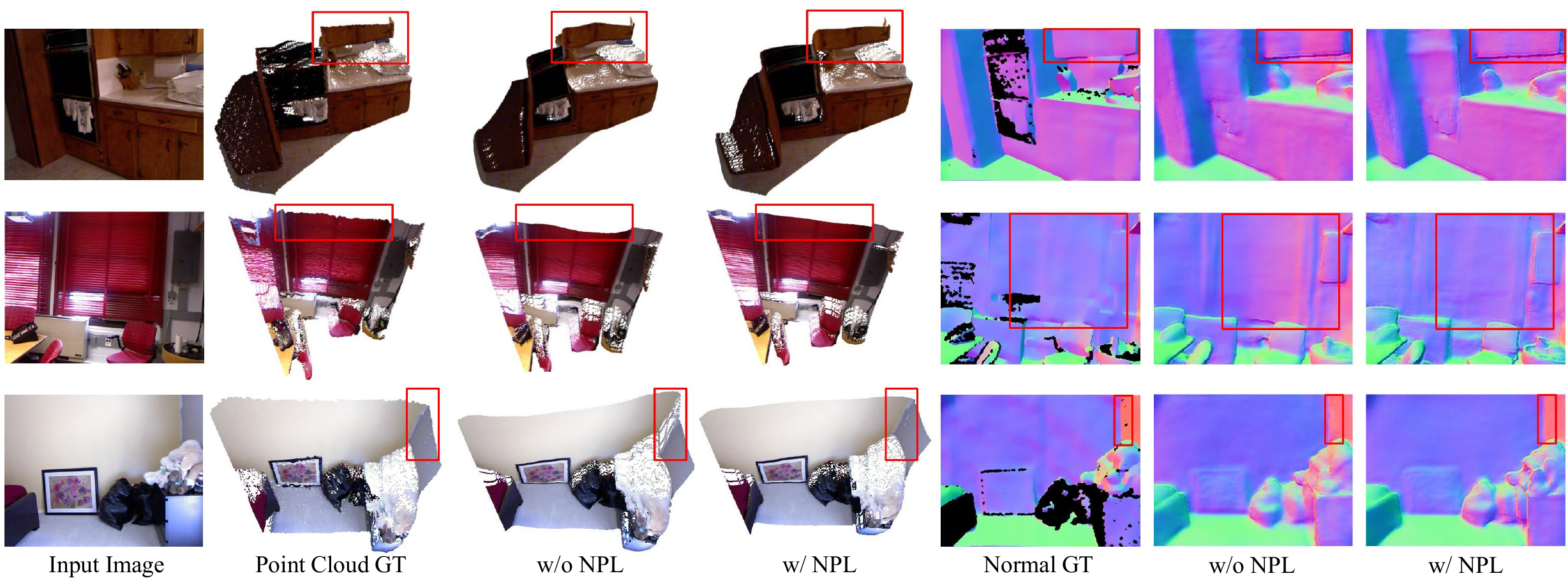}
    \caption{The visualization results of models with and without the normal prediction layer (NPL). With the normal prediction layer, the point clouds and recovered normal from the estimated depth are less distorted and keep more accurate structures. }
    \label{fig:pc_ablation}
    \vspace{-1mm}
\end{figure*}

\begin{figure*}[h]
\setlength{\abovecaptionskip}{1pt}
\setlength{\belowcaptionskip}{1pt}
    \centering
    \includegraphics[width=\linewidth]{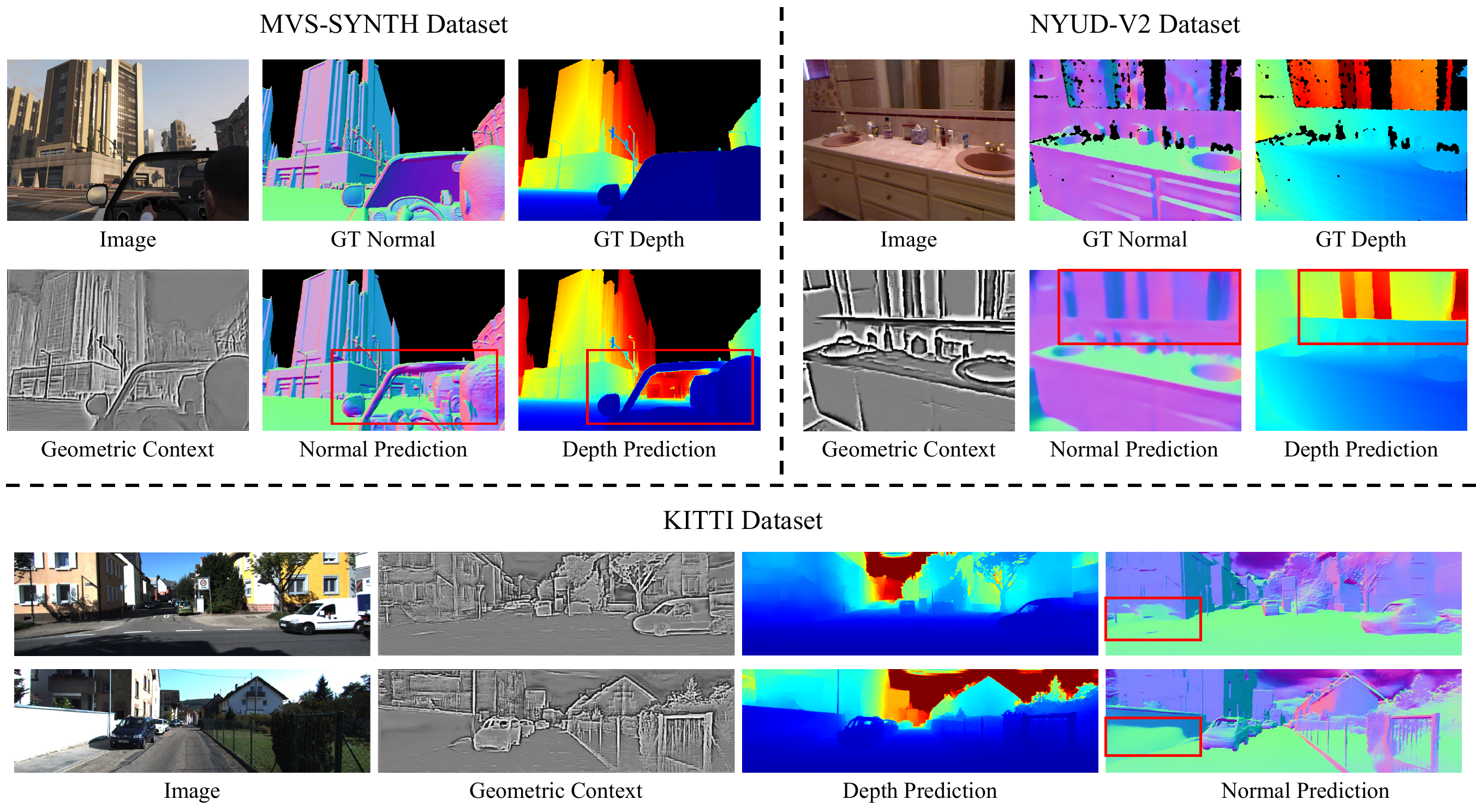}
    \caption{Failure cases of our method in indoor and outdoor datasets. In the MVS-SYNTH dataset, we show that our method fails to identify transparent objects while in the NYUD-V2 dataset, our method is unable to identify objects with specular reflections. We also test our method on the KITTI dataset where the model is only trained on the synthetic SEVERS dataset. It is evident that there is a noticeable performance drop.}
    \label{fig:failure}
    \vspace{-3mm}
\end{figure*}

\subsection{Discussions about normal estimation.}
In this part, we conduct experiments to validate the effectiveness of our proposed modules for normal estimation. Note that DPT is used as the backbone in our model for the experiments below.

\subsubsection{Effectiveness of guidance of geometric context }
To validate the effectiveness of our proposed geometric context guided normal estimation, we conduct experiments with the following settings: 
1) without guidance; 2) with guidance obtained from the first-order derivatives of geometric context; 3) with guidance obtained from the second-order derivatives of geometric context. The aforementioned settings are denoted as "w/o guidance", "1st-order", and "2nd-order" respectively. 
As shown in Table~\ref{tab:refinement}, the guidance of geometric context further improves the accuracy of the estimated normals. 
Using first-order or second-order derivatives to calculate the guidance weight maps achieve similar performance.
Although the usage of guidance can not lead to significant numerical improvement in Table~\ref{tab:refinement}, the predicted normals indeed exhibit improved geometric details at shape boundaries with the assistance of guidance, demonstrated in Fig.~\ref{fig:normal_ablation} (marked by red bounding boxes). Thus, we infer that the guidance of geometric context mainly benefits the regions with rich geometric details, which account for only a small portion of the whole image, resulting in an insignificantly quantitative improvement.

\subsubsection{Performance improvement by the normal decoder}
Although the normal decoder is used to predict normal maps via neural layers,  
jointly training normal estimation as a complementary task can improve the 3D geometric quality of the estimated depth. 
Taskonomy \cite{zamir2018taskonomy} has also demonstrated that joint learning between the tasks that are correlated can improve each task's performance.
We conduct extensive experiments to analyze the effects of the extra normal prediction layer. (For convenience,  we use \textbf{NPL} to represent the normal prediction layer in the following discussions.)
As shown in the Table~\ref{tab:guidance_normal_decoder}, including normal estimation as a complementary task significantly improves the geometry quality of the estimated depth, where the recovered normal and point clouds achieve higher accuracy.
We also visualize the recovered normals and point clouds generated by estimated depth in Fig.~\ref{fig:pc_ablation}. As marked in red bounding boxes, with the extra normal prediction layer, the point clouds converted from estimated depth preserve more accurate geometry and their corresponding normal maps contain less errors.

\vspace{-4mm}
\section{Conclusion and Future Works}
\label{clu}
In this paper, we present a simple but effective Adaptive Surface Normal (ASN) constraint to capture reliable geometric context, utilized to jointly estimate depth and surface normal with high quality. 
With the guidance of geometric context, our estimated depth maps and normal maps are robust to shape variations and can capture rich detailed geometries. Furthermore, our method faithfully preserves important 3D geometric features so that high-quality 3D point clouds and surface normal maps can be recovered from the estimated depth maps.
 Though our method successfully captures the geometric context in most cases, it is unable to identify objects with specular reflections or transparency. 
Since our model is only trained on synthetic data of outdoor scenes, our model suffers from a performance drop on the real outdoor scenes, as shown in Fig.~\ref{fig:failure} . In the future, we plan to incorporate segmentation methods for transparent and reflective surfaces to tackle the problem of estimating geometry for transparent and reflective surfaces. We also consider sim2real strategies to fully utilize synthetic data to boost the generalizability of our model on real outdoor scenes.

\begin{IEEEbiography}[{\includegraphics[width=1in,height=1.25in,clip,keepaspectratio]{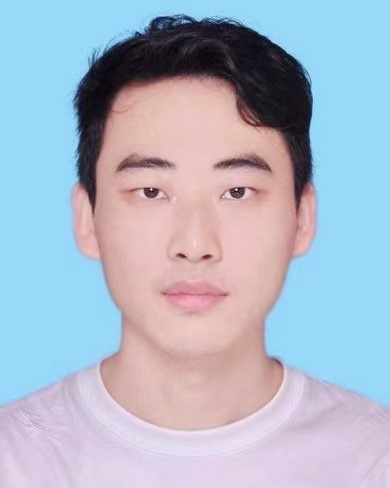}}]{Xiaoxiao Long}
received his Ph.D. from The University of Hong Kong (HKU), advised by Prof. Wenping Wang. Before that, he did his bachelor's degree in Control Science \& Engineering at Zhejiang University, supervised by Prof. Junfeng Wu. His research interests include 3D vision, depth estimation, and computer graphics. 
\end{IEEEbiography}

\begin{IEEEbiography}[{\includegraphics[width=1in,height=1.25in,clip,keepaspectratio]{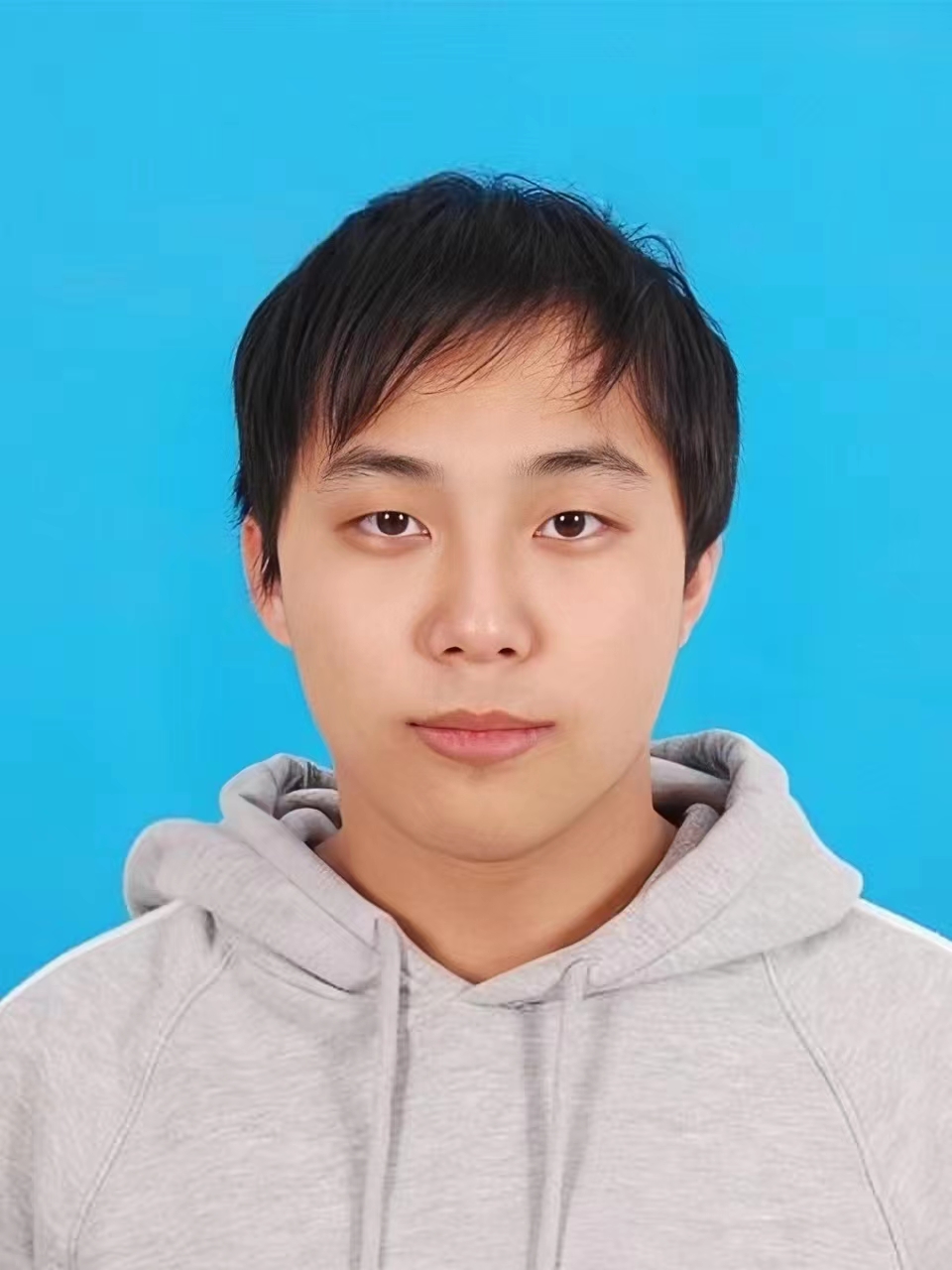}}]{Yuhang Zheng}
is currently an intern at Institute for AI Industry Research, Tsinghua University. His research interests include 3D scene understanding and robotics manipulation.
\end{IEEEbiography}

\begin{IEEEbiography}[{\includegraphics[width=1in,height=1.25in,clip,keepaspectratio]{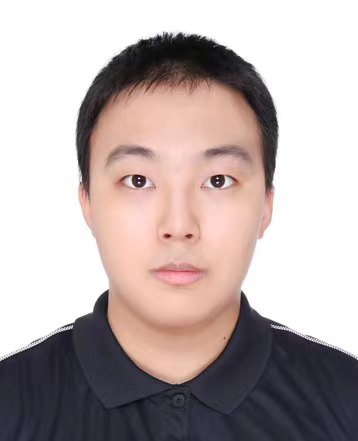}}]{Yupeng Zheng}
is currently an intern at Institute for AI Industry Research, Tsinghua University. His research interests include automatic driving and night-time depth estimation.
\end{IEEEbiography}

\begin{IEEEbiography}[{\includegraphics[width=1in,height=1.25in,clip,keepaspectratio]{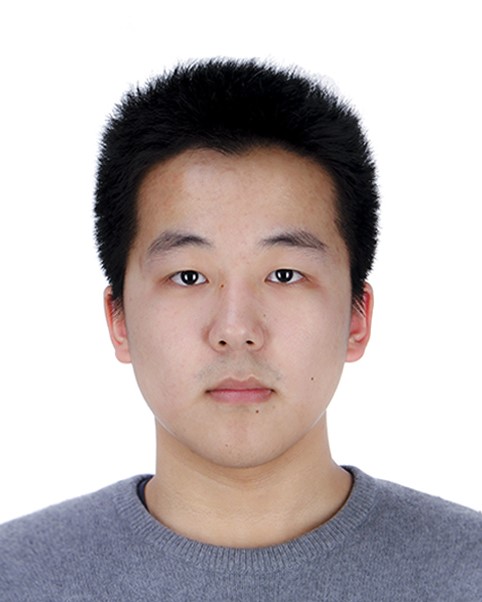}}]{Beiwen Tian}
is a Ph.D candidate at AIR, Tsinghua University, advised by Prof. Ya-Qin Zhang. Before that, he did his bachelor’s degree at Department of Computer Science and Technology, Tsinghua University. His research interests include 3D scene understanding, style transfer and weakly-supervised learning.
\end{IEEEbiography}

\begin{IEEEbiography}[{\includegraphics[width=1in,height=1.25in,clip,keepaspectratio]{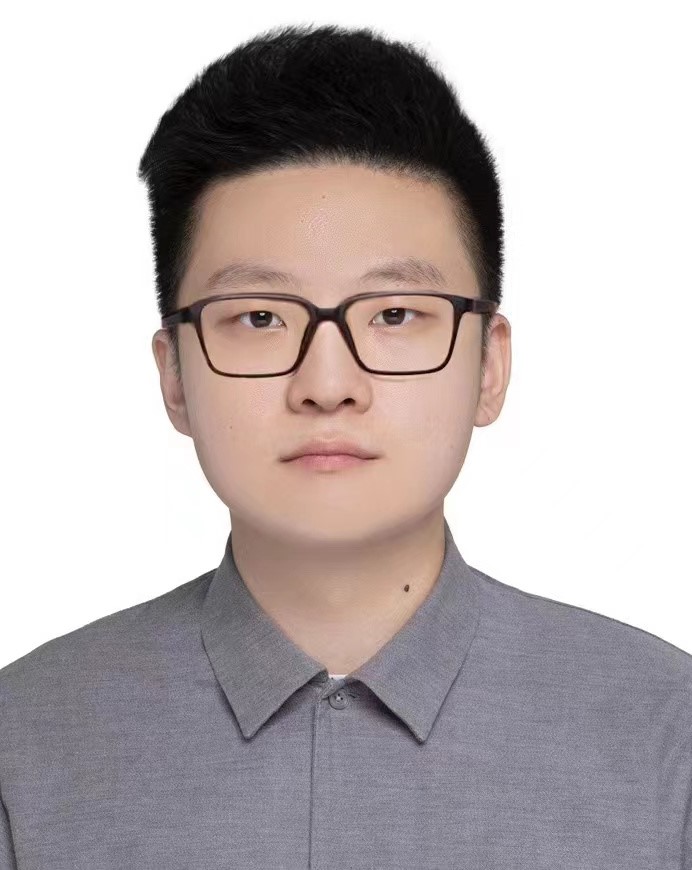}}]{Cheng Lin}
received his Ph.D. from The University of Hong Kong (HKU), advised by Prof. Wenping Wang. He visited the Visual Computing Group at Technical University of Munich (TUM), advised by Prof. Matthias Nießner. Before that, he did his bachelor's degree at Shandong University and was a Research Asistant at the Research Center of Digital Media Technology. His research interests include geometric modeling, 3D vision, shape analysis, and computer graphics.
\end{IEEEbiography}

\begin{IEEEbiography}[{\includegraphics[width=1in,height=1.25in,clip,keepaspectratio]{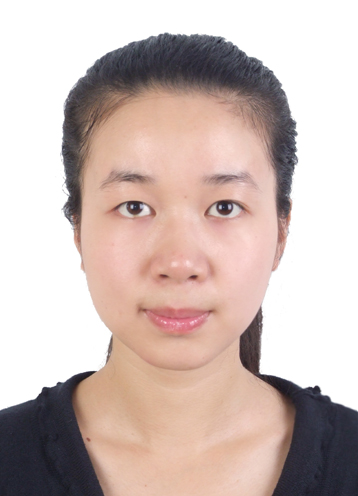}}]{Lingjie Liu}
is the Aravind K. Joshi Assistant Professor in the Department of Computer and Information Science at the University of Pennsylvania, where she leads the Penn Computer Graphics Lab and she is also a member of the General Robotics, Automation, Sensing \& Perception (GRASP) Lab. She received a B.E. degree from the Huazhong University of Science and Technology in 2014 and a Ph.D. degree from the University of Hong Kong in 2019. Her research interests include 3D reconstruction, human performance capture, and video synthesis. She has received Hong Kong Ph.D. Fellowship Award (2014) and Lise Meitner Fellowship Award (2019).
\end{IEEEbiography}

\begin{IEEEbiographynophoto}{Hao Zhao}
received the B.E. degree and the Ph.D. degree both from the EE department of Tsinghua University, Beijing, China. He is currently an Assistant Professor with the Institute for AI Industry Research (AIR), Tsinghua University. He was a research scientist at Intel Labs China and a joint postdoc affiliated to Peking University. His research interests cover various computer vision topics related to robotics, especially 3D scene understanding. Photograph not available at the time of publication.
\end{IEEEbiographynophoto}

\begin{IEEEbiographynophoto}{Guyue Zhou}
received the Ph.D. degree from the Hong Kong University of Science and Technology, Hong Kong, China, in 2014. He is currently an Associate Professor with the Institute for AI Industry Research (AIR), Tsinghua University. His research interests include advanced manufacturing, robotics, computer vision, and human-machine interaction. 
\end{IEEEbiographynophoto}

\begin{IEEEbiography}[{\includegraphics[width=1in,height=1.25in,clip,keepaspectratio]{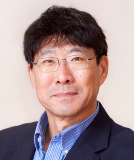}}]{Wenping Wang}
 is currently a Professor of the Department of Computer Science \& Engineering at Texas A\&M University. His research interests include computer graphics, visualization, and geometric computing. He has made fundamental research contributions in collision detection, shape modeling and analysis, mesh generation, and architectural geometry. He is the journal associate editor of Computer Aided Geometric Design (CAGD), Computers and Graphics (CAG), IEEE Transactions on Visualization and Computer Graphics (TVCG, 2008-2012), Computer Graphics Forum (CGF), IEEE Computer Graphics and Applications, and IEEE Transactions on Computers. He received the Outstanding Researcher Award of the University of Hong Kong in 2013. He received John Gregory Award in 2017 for contributions in geometric modeling and computing. He is an IEEE Fellow.
\end{IEEEbiography}






\clearpage
\section{Appendix}
\subsection{Choice of window size }
The choice of window size will influence the accuracy of recovered normal via ASN and its time consumption.
Generally speaking, the window size cannot be too small or too large according to how large the region covered by one image pixel is in the 3D world. 
If too small, the approximated normal of the window will be sensitive to noises. If too big, the local assumption will be not satisfied, causing inaccurate normal approximation.
We conduct experiments to analyze the choices of window size and found that there exists an optimal ratio that can be used to determine a suitable window size for different image resolutions.

\begin{figure*}[h]
\setlength{\abovecaptionskip}{1pt}
\setlength{\belowcaptionskip}{1pt}
    \centering
    \includegraphics[width=0.9\linewidth]{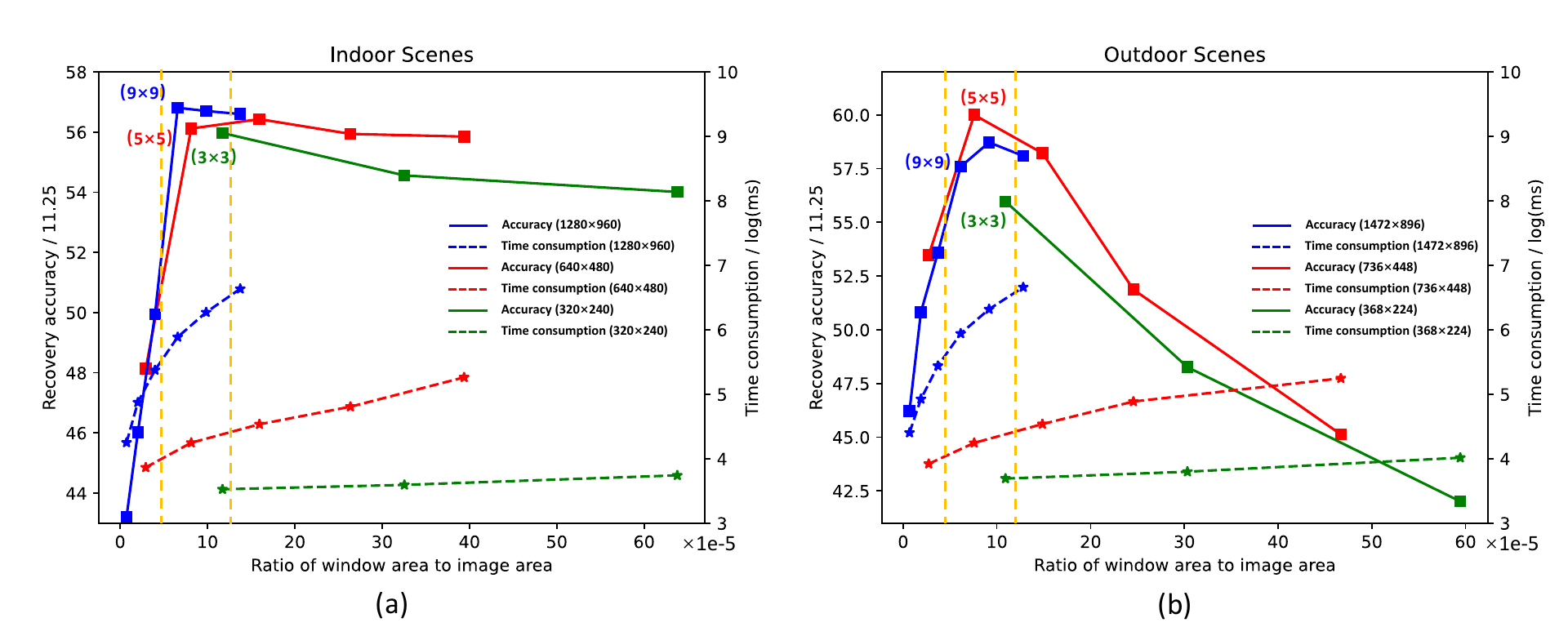}
    \caption{The analysis of accuracy and time consumption about recovered normal calculation with respect to window sizes of ASN constraint. 
    The X-axis represents the ratio of the window area to image area, with a scale of $1e^{-5}$. 
    The right Y-axis represents the time consumption of recovered normal calculation, which is logarithmically scaled milliseconds (log(ms)). 
    The left Y-axis represents the accuracy of the recovered normal maps, which is measured by the metric $\boldsymbol{11.25^{\circ}}$.
    The solid line represents normal accuracy, while the dashed line represents time consumption. 
    The colors blue, red and green represent the resolutions of the input image, high resolution, medium resolution and low resolution.
    Based on the experiments on indoor or outdoor scenes, considering accuracy and efficiency, the optimal window size for difference image resolution can be set by such rule: the ratio of window area to image area is around $5e^{-5} \sim 1e^{-4} $.}
    \label{fig:kernel}
    \vspace{-2mm}
\end{figure*}

As shown in Fig.~\ref{fig:kernel}, we analyze how the ratio of window area to image area influences the accuracy and time consumption of the recovered normal via ASN constraint.
For images with high/medium/low resolutions in indoor or outdoor scenes, the optimal ratio is around $5e^{-5} \sim 1e^{-4} $ (the orange dotted intervals in Fig.~\ref{fig:kernel}), since it leads to high accuracy and relatively low time consumption.
As a result, we can leverage the optimal ratio to determine a window size for a given image resolution.
Taking the image resolution of $1280\times960$ for example, when we assume the ratio is $8e^{-5}$ the window size should be $ \sqrt{1280\times960 \times 8e^{-5}}=9.9$. Since the window size should be odd integer, and it can be set to $9 \times 9$.

\subsection{More qualitative results of our method}
\begin{figure*}[!t]
\setlength{\abovecaptionskip}{1pt}
\setlength{\belowcaptionskip}{1pt}
    \centering
    \includegraphics[width=\linewidth]{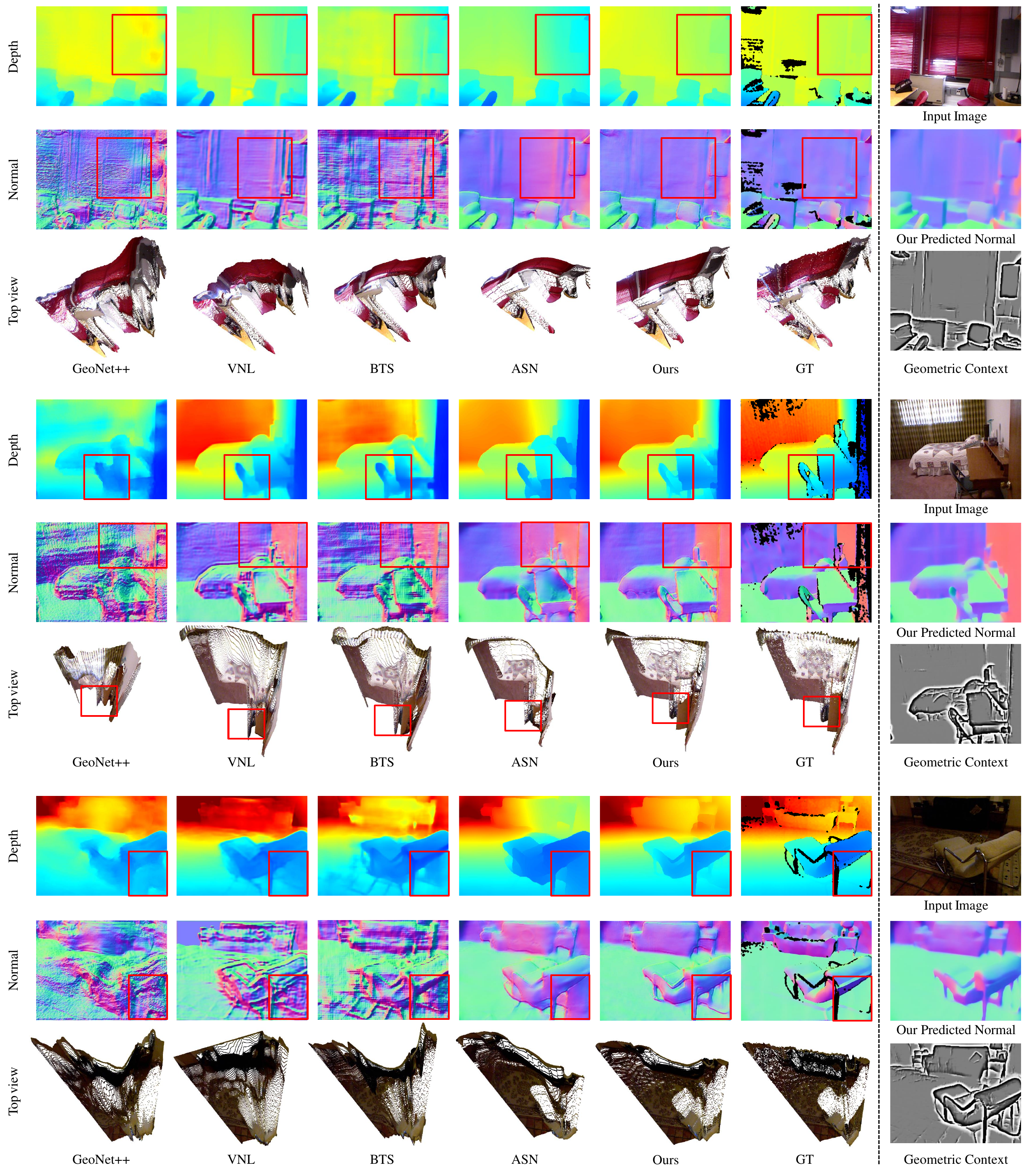}
    \caption{Qualitative comparisons on NYUD-V2 dataset. We compare our predicted depth, recovered surface normals, and point clouds with other SOTAs. We also show our predicted normal and geometric context.  }
    \label{fig:nyu_sup}
    \vspace{-3mm}
\end{figure*}

\begin{figure*}[!t]
\setlength{\abovecaptionskip}{1pt}
\setlength{\belowcaptionskip}{1pt}
    \centering
    \includegraphics[width=\linewidth]{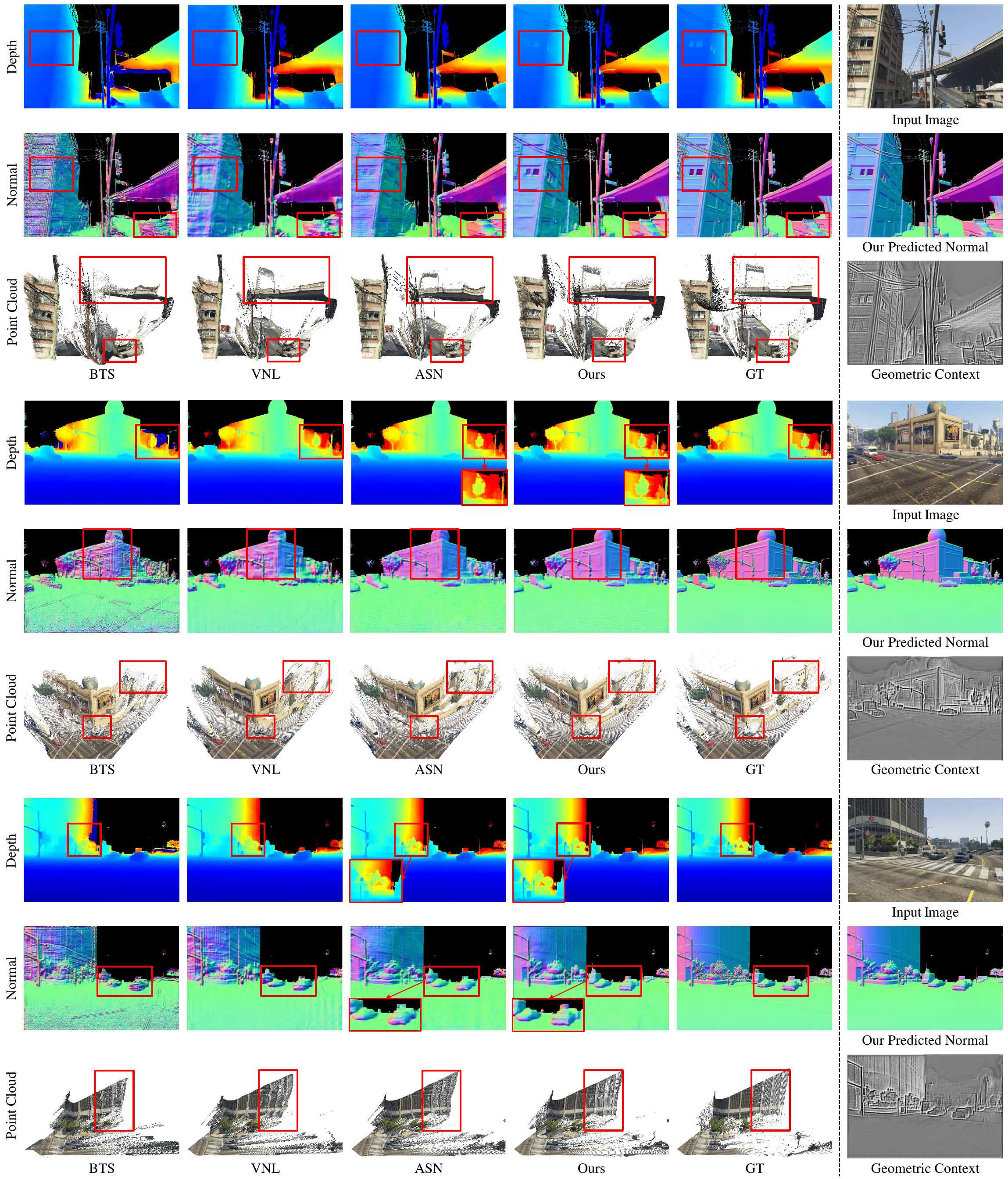}
    \caption{Qualitative comparisons on MVS-SYNTH dataset. We compare our predicted depth, recovered surface normals, and point clouds with other SOTAs. We also show our predicted normal and geometric context. }
    \label{fig:mvs_sup}
    \vspace{-3mm}
\end{figure*}

\begin{figure*}[!t]
\setlength{\abovecaptionskip}{1pt}
\setlength{\belowcaptionskip}{1pt}
    \centering
    \includegraphics[width=\linewidth]{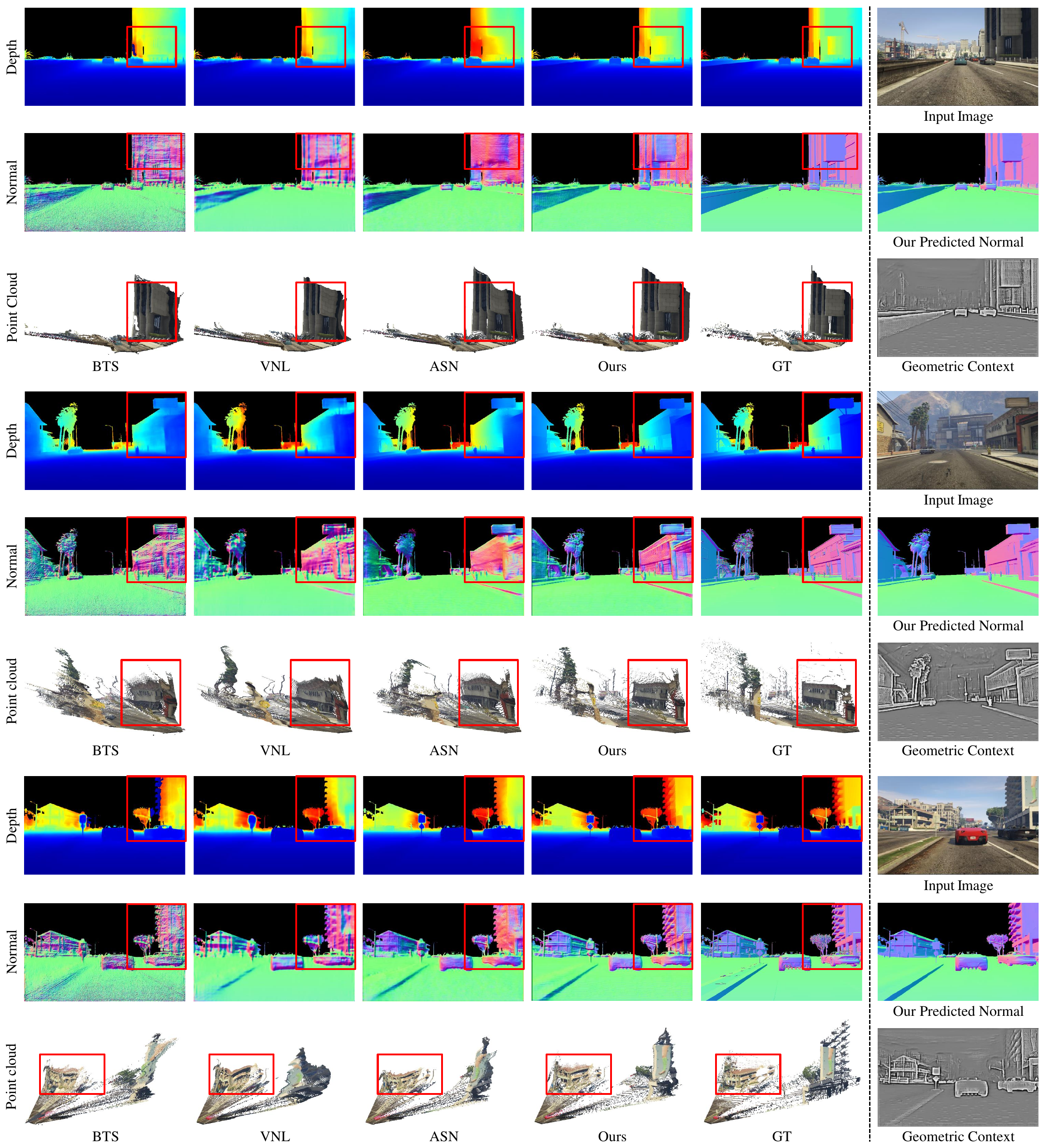}
    \caption{Qualitative comparisons on SVERS dataset. We compare our predicted depth, recovered surface normals, and point clouds with other SOTAs. We also show our predicted normal and geometric context.  }
    \label{fig:svers_sup}
    \vspace{-3mm}
\end{figure*}

We show more qualitative results on NYUD-V2 dataset, MVS-SYNTH dataset, and SVERS dataset, as depicted in Fig.~\ref{fig:nyu_sup}, Fig.~\ref{fig:mvs_sup} and Fig.~\ref{fig:svers_sup} respectively. 

We compare our predicted depth, recovered surface normals, and reconstructed point clouds with other SOTAs. The results prove that our method performs better on regions with geometric variation in both indoor scenes and outdoor scenes. The detailed comparisons can be seen in red bounding boxes. We also show our predicted surface normal. It can be seen that our predicted normals achieve high quality.
Besides, based on the geometric context, our method has a strong ability to capture geometric variations and rich detailed geometries.  

\subsection{More qualitative results of guidance module}
\begin{figure*}[t]
\setlength{\abovecaptionskip}{1pt}
\setlength{\belowcaptionskip}{1pt}
    \centering
    \includegraphics[width=\linewidth]{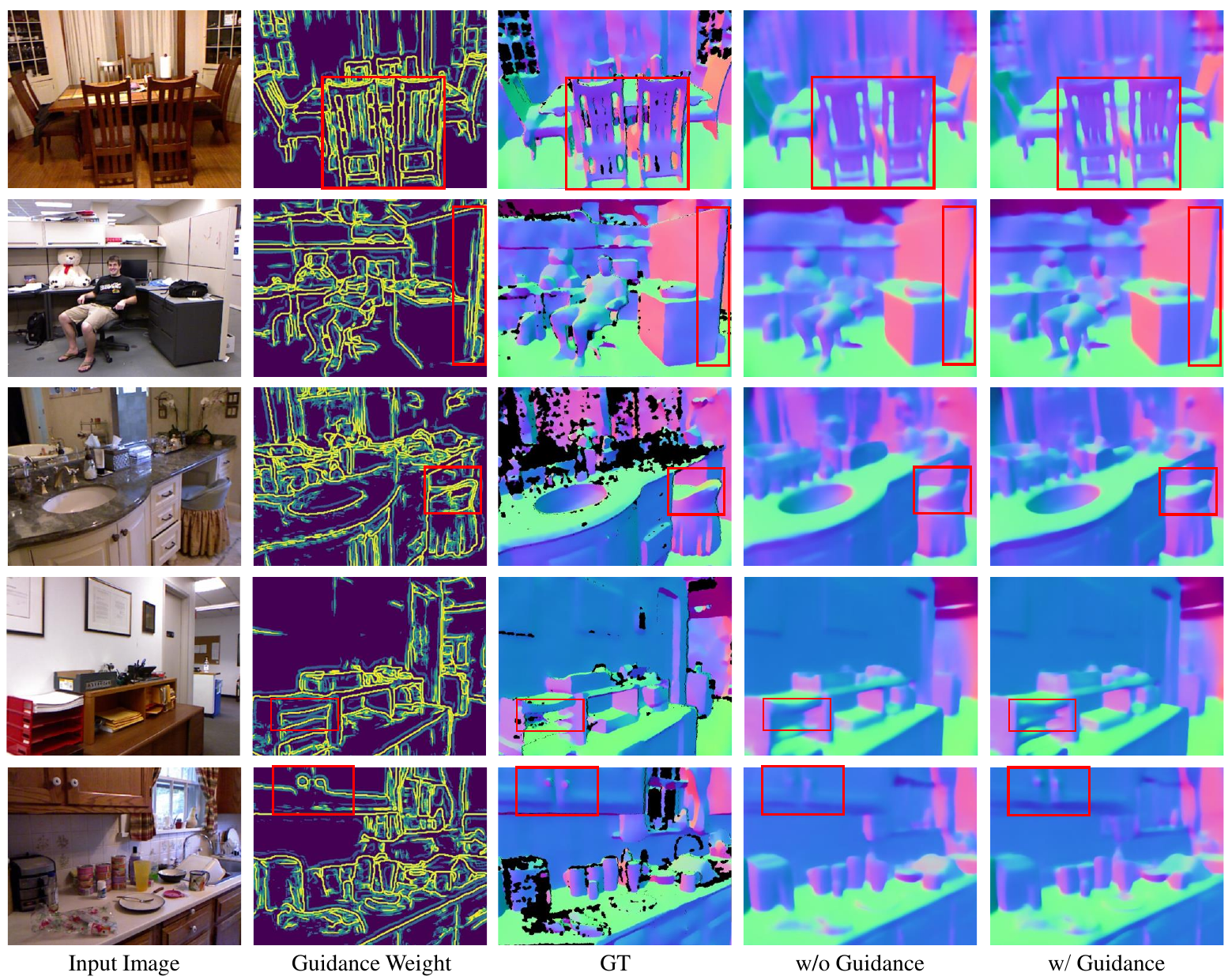}
    \caption{Qualitative comparisons of our method with and without the guidance of geometric context. The guidance of geometric context indeed boosts the ability to capture detailed geometries (red bounding boxes).
    }
    \label{fig:ab_normal_sup}
\end{figure*}

We show more qualitative results with and without the guidance of geometric context to prove the effectiveness of the proposed approach for normal estimation. As shown in Fig.~\ref{fig:ab_normal_sup}, it can be seen that with the guidance of geometric context, our predicted surface normals are more accurate on regions with rich geometric details, such as boundaries and small structures.

\end{document}